\documentclass{article}

\usepackage{arxiv}

\usepackage[utf8]{inputenc} 
\usepackage[T1]{fontenc}    
\usepackage{hyperref}       
\usepackage{url}            
\usepackage{booktabs}       
\usepackage{amsfonts}       
\usepackage{nicefrac}       
\usepackage{microtype}      
\usepackage{lipsum}
\usepackage{graphicx}
\usepackage{subcaption}
\usepackage{longtable}
\usepackage{amsmath}
\usepackage[title]{appendix}
\usepackage[square, numbers]{natbib}
\graphicspath{ {./images/} }

\title{Transformers Meet Hyperspectral Imaging: A Comprehensive Study of Models, Challenges and Open Problems}

\author{
 Guyang Zhang \\
  Department of Electrical, Computer and Software Engineering\\
  University of Auckland\\
  5 Grafton Rd, Auckland Central, Auckland 1010, New Zealand \\
  \texttt{gzha422@aucklanduni.ac.nz} \\
   \And
 Waleed Abdulla \\
  Department of Electrical, Computer and Software Engineering\\
  University of Auckland\\
  5 Grafton Rd, Auckland Central, Auckland 1010, New Zealand \\
  \texttt{w.abdulla@auckland.ac.nz} \\  
}

\begin{document}
\maketitle
\begin{abstract}
Transformers have become the architecture of choice for learning long-range dependencies, yet their adoption in hyperspectral imaging (HSI) is still emerging. We reviewed more than 300 papers published up to 2025 and present the first end-to-end survey dedicated to Transformer-based HSI classification. The study categorizes every stage of a typical pipeline—pre-processing, patch or pixel tokenization, positional encoding, spatial-spectral feature extraction, multi-head self-attention variants, skip connections, and loss design—and contrasts alternative design choices with the unique spatial-spectral properties of HSI. We map the field’s progress against persistent obstacles: scarce labeled data, extreme spectral dimensionality, computational overhead, and limited model explainability. Finally, we outline a research agenda prioritizing valuable public data sets, lightweight on-edge models, illumination and sensor shifts robustness, and intrinsically interpretable attention mechanisms. Our goal is to guide researchers in selecting, combining, or extending Transformer components that are truly fit for purpose for next-generation HSI applications.
\end{abstract}

\keywords{hyperspectral imaging \and Transformers \and deep learning \and classification \and neural networks}

\section{Introduction}
	
	Hyperspectral imaging (HSI) is an advanced imaging technique that simultaneously captures spatial and rich spectral information. Unlike RGB imaging, which only captures three visible colors, HSI can record the electromagnetic energy from the visible to infrared bands, which capture the target objects' reflection or absorption characteristics of specific wavelengths. HSI also extends the capabilities of spectrometers by providing the spectral information of the entire scanning area rather than a single point. 
	
	The data captured by the HSI system is a three-dimensional datacube, which combines one-dimensional spectral features and two-dimensional spatial information. The broad spectrum range and abundant spatial-spectral information allow HSI to distinguish objects and materials more accurately by identifying properties invisible to the naked eye \cite{Yun2023SpecTr}. Moreover, the non-destructive, non-invasive, fast, and simple properties render HSI appropriate for various applications in multiple areas, including remote sensing \cite{KHAN2022101678}, fruit and vegetables quality assessment \cite{WIEME2022156}, meat quality assessment \cite{JO2024109785}, stress and contamination detections of agriculture products \cite{BARBEDO2023107920}, forensic science \cite{MARIOTTI2023387}, medical image analysis \cite{REHMAN2021102165}, and various scientific areas \cite{Kazi2023hsiscience}.  
	
	With the fast development of deep learning, various algorithms have been applied with the HSI technique \cite{KHAN2022101678}, such as convolutional neural networks (CNN) \cite{YU2023108382, NOSHIRI2023100316}, recurrent neural network \cite{Paoletti2020}, autoencoder \cite{JAISWAL2023100584}, and Transformer \cite{Perera2023lowpixel, Lin2023metasurface}. Due to the unique data characteristics of HSI, fully exploring and emphasizing the essential information contained in HSI data is challenging \cite{Yang2023barrier}. Thus, algorithms processing HSI data should sufficiently combine the spatial-spectral information rather than focusing excessively on spatial or spectral features \cite{Yang2023gtfn}. Among these algorithms, CNN-based methods, which are restricted by a limited receptive field, are not good at modeling the long-range dependencies and are challenging to explore and represent the sequence attributes of spectral signatures \cite{He2020hsibert, rs13030498, Hong2022spectralformer, Yang2022hsiT, Sun2022tokenization, Shi2023fcanakf}. The 3-D convolution, which can process data cubes along all three dimensions, incurs high computational complexity \cite{Gao2023MSTSSSNet}.  
	
	In contrast, the Transformer architecture, which was initially designed for natural language processing (NLP) tasks \cite{Vaswani2017attention}, can capture the long-range context relationships among input tokens through the multi-head self-attention mechanism (MHSA) and demonstrates capabilities for computer vision (CV) tasks with representative works including Vision Transformer (ViT) \cite{dosovitskiy2021an} and Swin Transformer (SwinT) \cite{Liu2021swint}. Numerous CV tasks have adopted Transformer-based networks for applications in different areas, such as plant disease identification \cite{YANG2023107809}, crop mapping \cite{LI2022107497}, and weed identification \cite{ESPEJOGARCIA2023108055}. 
	
	Because the non-adjacent spectral bands of HSI data display long-term dependency with each other \cite{ZHOU201939lstm}, the Transformer is suitable to process the spatial-spectral information of HSI data. Therefore, Transformer-based models have been widely explored to be applied to remote sensing tasks using HSI data \cite{Yuan2021pretrainvit, Liu2022central, AZAD2024103000}, such as oil spill mapping \cite{Kang2023ssloil}, methane detection \cite{Kumar2023methanemapper}, crop mapping from UAV \cite{NIU2022107297}, grassland degradation monitoring using UAV \cite{Zhang18032024}, crop stress classification \cite{KHOTIMAH2023103286}, crop field mosaic generation \cite{Perera2023lowpixel}, and forestry tree species classification \cite{Zhang2023morpho}. Transformer-based models have also been applied to proximal analyzing tasks, including cancer areas segmentation \cite{Zhou2021swinspectral, Yun2023SpecTr}, in-vivo brain tumor tissue detection \cite{Cruz-Guerrero2023braintissues, Sigger2024brain}, blood cell classification \cite{Li2023shift}, maize seed variety recognition \cite{agronomy12081843, Domezz2023hybrid}, concrete cracks segmentation \cite{Steiner2023concrete}, HSI video object tracking \cite{Gao2023cbffNet}, point cloud segmentation \cite{Afifi2023tinto}, soluble solid content and pH prediction of cherry tomatoes \cite{foods13020251}, and infectious bacteria identification \cite{Lu2024bacteria}.
	
	Although Transformer-based models can better process long-range dependencies, they are prone to ignoring some local information. The local information is important for HSI classification to generate accurate token embeddings from a single dimension of HSI cubes \cite{Zhao2022contrans, Zou2022LessFormer}. In addition, Transformer-based models usually require sufficiently large training samples, and their performance relies heavily on large-scale pretraining, but HSI suffers from limited annotated samples \cite{Zhou2022hussat, Zhou2023ViTContrast, Mohamed2023factorformer}. The high-dimensional spectral data of HSI, which contains redundant information, leads to a heavy memory burden and huge computational cost, together with the MHSA mechanism \cite{Zhang2023lightweight}, and hinders the interaction between features from distant locations \cite{liang2022hsimixer, Sun2023largekernel}. 
	
	However, it is challenging to balance the trade-off between model complexity and generalizability because reducing model complexity is beneficial to alleviate overfitting, whereas utilizing more complex models helps enhance feature extraction \cite{Xiao2023dsdcdcam}. Therefore, many research works have been devoted to addressing these challenges, and diversified techniques have been proposed from different perspectives to improve the performance of Transformer-based models with HSI tasks.
	
	Nowadays, many authors have already investigated the Transformer from various perspectives, such as applications on deep learning tasks \cite{ISLAM2024122666}, Transformers in computer vision tasks \cite{Khan2022tiv, XuYifan2022Ticv, Han2023vit}, architecture modification \cite{LIN2022111}, improving model efficiency and optimization \cite{Tay2020EfficientTA, CHITTYVENKATA2023102990, Fournier2023practical}, neural architecture search \cite{Krishna2022NAS}, text summation \cite{Guan2021text}, visualizing Transformers for NLP \cite{Brasoveanu2020}, medical imaging analysis \cite{AZAD2024103000, LI2023102762, SHAMSHAD2023102802, XIAO2023104791, LIU2023107268, WANG2023100004}, multimodal learning \cite{Xu2023multimodalTr}, object tracking \cite{Kugarajeevan2023obtrack}, Transformers for video tasks \cite{Selva2023videoT}, and Transformers in Remote Sensing \cite{rs15071860review, s24113495rsTreview}. 
	
	There are also a large number of reviews about HSI, including research trends of HSI analysis \cite{Khan2018hsitrend}, HSI classification and prediction \cite{Tejasree2024review}, proximal HSI in agriculture applications \cite{BARBEDO2023107920}, HSI unmixing \cite{Bhatt2020unmix, Heylen2014nonlinunmix}, object tracking using hyperspectral videos \cite{Qian2023survey}, comparing the performance of different algorithms on HSI \cite{Viel2023analysis}, HSI Remote Sensing Classification with UAV \cite{Zhang2025uav}, integration of natural language processing (NLP) techniques with HSI \cite{Akewar2025LLM}, and HSI classification with limited samples \cite{Ullah2025survey}. There are surveys covering deep learning methods on HSI classification \cite{KUMAR2024100658, HAIDARH2025}, and reviewing all algorithms for HSI classification from conventional methods to the most advanced Transformer and Mamba \cite{ahmad2024survey}.  
	
	To the best of our knowledge, we have not found any reviews focusing on the application of Transformer-based models on the HSI classification task. There is one work that investigated the HSI classification based on graph neural networks (GNN) \cite{Zhao2025reviewgnn} and another work about the application of autoencoder (AE) on HSI tasks \cite{JAISWAL2023100584}, which briefly reviewed how AE can benefit the HSI tasks and applications of AE on different tasks, including classification, unmixing, and anomaly detection. 
	
	This article aims to present a comprehensive survey on the application of Transformer-based networks on hyperspectral imaging classification tasks. Considering the large amount of research work on integrating Transformers with HSI data, other computer vision tasks with hyperspectral imaging, such as target detection \cite{Zheng2022ood, Qin2022HTD-Vit, Feng2023attentionmulti}, anomaly detection \cite{Xiao2023anomaly, Li2023randommasks, He2023CTA}, change detection \cite{Wang2022sst, Dong2023graphT, Ding2022CDFormer}, denoising \cite{Chen2022Hider, Ibanez2022MAE, Wang2023VariationLoss}, unmixing \cite{Ghosh2022unmixing, Kong2023sparseunmixing, Duan2023undat}, multimodal data fusion \cite{Xue2022sslmultimodal, Xue2022Hvit, Roy2023multimodal, Gao2023csm, Li2024swformer}, super-resolution \cite{Hu2022fusformer, Liu2022interactformer, Xu2023TransUNet}, hyperspectral image reconstruction \cite{Cai2022maskcvpr, Cai2022mst, Yu2023dstrans}, and neural architecture search \cite{Zhong2021factor, Xue2022NAS}, will not be included in this article. 
	
	To generalize current research methods and inspire future research directions, we collected and analyzed the most recent research papers published in journals and conferences on integrating Transformer and hyperspectral imaging classification until December 2024. In order to gather relevant papers, we searched keywords "Transformer + hyperspectral + classification" or "self-attention + hyperspectral + classification" in research databases, including "IEEE," "ScienceDirect," "SCOPUS," "MDPI," "Springer," "Wiley," "Taylor \& Francis," and "Sage." In addition, some works that did not contribute sufficiently novel modifications to HSI classification or Transformer architectures or did not use self-attention mechanisms were filtered \cite{liang2022hsimixer, rs15040983}. A summary of the papers categorized according to different research repositories is demonstrated in Table \ref{databases}. 
	
	This study is structured to summarize different phases of utilizing Transformer-based networks for HSI classification tasks. Section \ref{limited} briefly reviews different learning techniques to address the issue of limited HSI samples. Section \ref{prep} summarizes some commonly used preprocessing methods for HSI. Section \ref{embed} illustrates different methods of patch splitting for HSI and token/positional embedding for Transformers. Section \ref{featureextract} categorizes the feature extraction and feature fusion methods according to network architectures. Section \ref{mhsa} depicts various modifications to multi-head self-attention mechanisms. Section \ref{sc} generalizes skip-connection methods, and section \ref{loss} discusses loss functions commonly implemented in Transformers on HSI classification.
	
	\section{Classical system for Transformer-based HSI classification}
	
\begin{figure}
		\centering
		\includegraphics[scale=0.5]{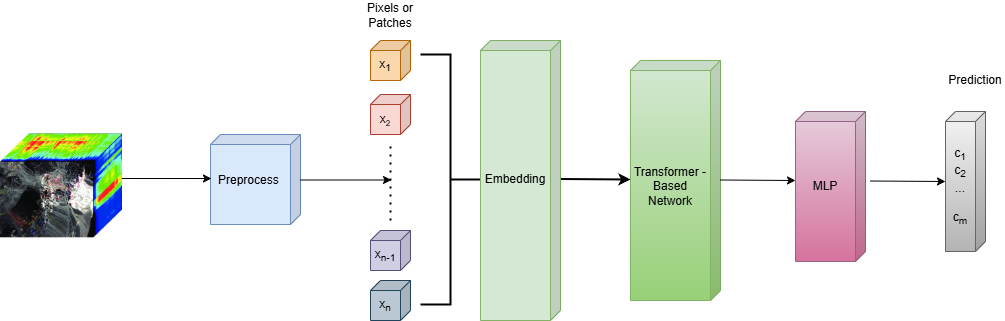}
		\caption{This figure shows a classical framework of a Transformer-based network on an HSI classification task, where the preprocessing module is optional. The hyperspectral image can be split into $n$ patches or pixels, denoted as $x_1, x_2, \dots x_{n-1}, x_n$. With $m$ different classes, the output predictions were denoted as $c_1, c_2, \dots, c_m$.}
		\label{classic}
	\end{figure}

	Figure \ref{classic} illustrates a typical framework for HSI classification with Transformer-based networks. Given an input hyperspectral image, it is optional to conduct pre-processing, such as Principal Component Analysis (PCA) or normalization, on the input image, while many works use the raw image \cite{He2020hsibert, rs13030498}. Then, the hyperspectral image can be split into patches or pixels and projected into the token embeddings \cite{He2020hsibert, Hong2022spectralformer} with positional embedding \cite{Chen2021multistageViT}. The Transformer-based networks take the token embeddings as input and output a feature map to be processed by an MLP module to generate the final prediction. The following sections will review the modifications to these different modules.
	
\begin{figure}
		\centering
		\includegraphics[scale=0.5]{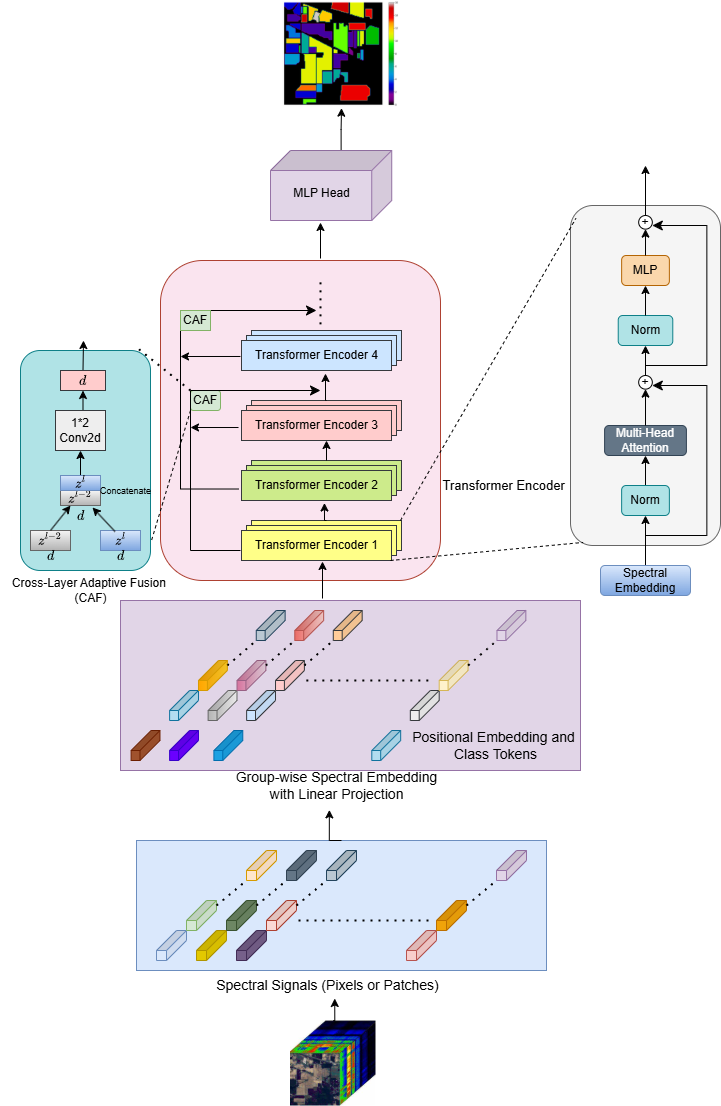}
		\caption{An overview of the framework of SpectralFormer \cite{Hong2022spectralformer}. }
		\label{spectralformer}
	\end{figure}

	For example, Fig. \ref{spectralformer} displays the framework of SpectralFormer \cite{Hong2022spectralformer}. The input HSI cube is first divided into pixels or image patches. Then, the spectral bands of these pixels or patches are partitioned into overlapping groups containing neighboring bands. The grouped spectral signals are linearly projected into embedding, added with positional embedding, and concatenated with class tokens, which are learnable parameters. Cascaded Transformer encoders subsequently process the spectral embeddings. Each Transformer encoder includes a Layer Norm layer, a multi-head self-attention module (MHSA), another Layer Norm layer, and a feedforward (denoted as MLP) module. 
	
	The input of the first Layer Norm is skip-connected with the output of the MHSA module, and this output feature is also skip-connected with the output of the MLP module. Moreover, the extracted features of these Transformer encoders are also skip-connected by Cross-Layer Adaptive Fusion (CAF) modules, which fuse the output features $z^l$ of layer $l$ and $z^{l-2}$ of layer $l-2$ with concatenation and 2D convolutional layer. The features generated by these Transformer encoders are then fed into the MLP head, which consists of Layer Norm and linear projection, to produce the classification prediction.
	
	\section{Limited Number of Samples Challenge}\label{limited}
	
	The Transformer models usually contain a large number of parameters. Therefore, the insufficient labeled samples of HSI lead to overfitting \cite{Li2023spformer}. However, labeled samples are limited for HSI classification since sensor differences, complex surface distribution, spatiotemporal heterogeneity, and atmospheric changes make manual labeling time-consuming, laborious, and costly to collect \cite{Huang2022swinvit, Huang2022swinvit, Yang2023iter}. The annotated samples are even more scarce in specific application areas, such as oil spilling detection \cite{Kang2023ssloil}. Therefore, various techniques have been proposed to address the issue of limited HSI samples.

	\begin{enumerate}
		\item \textbf{Transfer learning}: Many researchers have proposed addressing the issue of insufficient labeled samples by transferring information from a domain with abundant samples. The samples to be classified are from the target domain, and the datasets with sufficient samples for training are called the source domain. The cross-domain classification aims to exploit the similarities between the source and target domains to enhance the performance of the classifiers. 
		
		One way is to utilize off-the-shelf backbone networks, such as pre-trained VGGNet \cite{rs13030498, Wang2023DCN-T} and EfficientNet \cite{Domezz2023hybrid}, for feature extraction. However, the RGB dataset for pretraining of these backbones only has three channels, while the target HSI contains hundreds of channels. Therefore, \cite{rs13030498} adopted a mapping module that contains three learnable parameters to transform HSI patches into cubes with three channels. In addition, DCN-T \cite{Wang2023DCN-T} split the HSI channels along the spectral direction and aggregated the channels in each group by calculating the mean values of these channels. Three of these averaged sub-group values were selected to construct sub-images according to the preferences of different target objects. CMTL \cite{Cheng2023causalmeta} used principal component analysis (PCA) to unify the feature dimensions of source and target domains. 
		
		However, there is no guarantee that samples of source and target scenes are from identical distributions, resulting in significant performance degradation \cite{Zhao2022crossscene}, and the differences between the RGB source domain and HSI target domain are too significant. Thus, most other works in this area utilized HSI datasets as source and target domains and developed various methods to align features from different domains. For instance, FFTN \cite{Liu2022feedback} trained the classifier on the source domain, improved the distinguishing ability of misclassified classes by reinforced training (RT), and transferred the learned classifier to the target domain based on transductive learning (TL), which fine-tuned the classifier with small-sized labeled samples of the target domain. 
		
		Researchers also developed various techniques to align the features from the source domain and the target domain. ToMF-B \cite{Zhao2022crossscene} decomposed the extracted features into task-related and task-irrelevant (negative) features, which were determined by gradients of source classes from the view of channelwise attention. The domain alignment was performed in an adversarial manner: first, minimize the loss of labeled source data, then maximize the domain discrepancy between two independent classifiers with different initialization while minimizing the source error, and finally reduce the interdomain gap by updating the generator \cite{Zhao2022crossscene}. CMTL \cite{Cheng2023causalmeta} implemented a causal mask to segment each HSI into causal and noncausal regions, within which region pixels were substituted with randomly chosen heterogeneous pixels. The distribution differences between domains were also alleviated through an adversarial optimization process. Category prototypes were computed using the features of support samples, and the prediction of query samples was calculated by measuring the distance between the query feature and category prototypes in the feature space. 
		
		Moreover, Grid-Transformer \cite{Guo2023gridfewshot} mapped the source and target data into a shareable discriminative metric space, and CD-ViT \cite{Ling2023cdvit} aligned source and target features via the cross-attention module, which computed the query ($Q$) with source features and key ($K$) and value ($V$) with target features. CD-DViT \cite{Ye2024CD-DViT} designed a domain mapping (DM) branch to extract source and target features and then map the source features to the target domain—the mapped samples served as the target training samples. CD-DViT also implemented cross-attention, which used $Q$ from the source domain and $K$ and $V$ from the target domain. \cite{Ding2023crossdomain} proposed cross-domain calibration by transforming the source domain features more similar to Gaussian distribution and selecting statistics to calibrate the features in the target domain. In order to alleviate domain shift, CSJA \cite{Li2024csja} utilized the pseudo-labels from the target domain and shared representation features to predict samples from the source domain so that the pseudo-labels' quality in the source domain can be evaluated.
		
		Additionally, DT-FSL \cite{Ran2023fewshot} conducted classification on both source and target domains with only a few samples and realized distribution alignment as a minimax optimization problem, which minimized the loss metric of intra-domain discriminator (IDD, designed to discriminate whether the features are from source or target domains) and maximized the probabilities of correctly classifying source labels. CTFSL \cite{Peng2023crossdomain} also performed few-shot learning on both the source and target datasets concurrently. Then, it aligned the source and target domains into an identical dimension with a 2D CNN-based distribution aligner before the feature extractor. It distinguished the two domain classes by a domain discriminator to reduce domain shifts.

		\item \textbf{Self-supervised learning} is an unsupervised learning paradigm to tackle the challenge of insufficient labels \cite{Yuan2021pretrainvit}. The core idea of self-supervised learning is to pre-train the models without the labels and then fine-tune the pre-trained networks with a few labels in downstream tasks \cite{Huang2022swinvit}. Self-supervised learning methods can be categorized into two representative types: contrastive learning (CL) and masked autoencoder (MAE). CL aims to generate transferable visual representations by learning to be invariant to different data augmentations, and the MAE is designed to learn the representative features by masking and forcing the model to reconstruct the image \cite{oord2019representation, He2022cvpr}. 
		
		Many works have implemented MAE using different approaches for spatial-spectral feature masking. For example, SITS-BERT \cite{Yuan2021pretrainvit} pre-trained the encoder to predict the contaminated observations by randomly adding noise from a uniform distribution. MSBMSW \cite{Li2023shift} masked spectral bands with a fixed ratio and then designed spectral-wise U-Net to reconstruct the missing spectral bands using mean squared error (MSE) loss. In the fine-tuning phase, the encoder, which contains the pre-trained parameters, is connected to a classifier. SSLSM \cite{Liu2023sslmask} also adopted limited spectral masking while preserving the spatial information and reconstructed masked data with the Transformer-based decoder. 
		
		MSSFP \cite{Zhou2023MSSFP} randomly masked a portion of two sequences extracted from spectral and spatial dimensions, and the Transformer is trained to predict the principal components (PCs) and local binary pattern (LBP) of masked regions. MaskedSST \cite{Linus2023marsked} randomly masked HSI spatial-spectral patches, then reconstructed the pixel values, which were measured by L1 loss, using a small linear mapping layer. FactoFormer \cite{Mohamed2023factorformer} also used simple linear layers as decoders to reconstruct masked regions but with MSE as the loss function. 
		
		IMAE \cite{Kong2023instoken} utilized random tokens to replace the masked embeddings and Transformer-based encoder, which was also adopted by SS-MTr \cite{Huang2023markedtrans}, to reconstruct the original HSI patches. However, SS-MTr can be extended to discriminative learning tasks, including contrastive learning, supervised learning, and supervised-contrastive learning, by adding extra branches for specific tasks in parallel with the pre-trained networks \cite{Huang2023markedtrans}. In addition, adaptive channel module and masked self-supervised pretraining were adopted to train models across multiple HSI datasets to solve the challenges of band and label differences \cite{Bai2024crossdata}. RMAE \cite{Wang2024RMAE} proposed a regularized MAE with parallel Transformer branches to reconstruct 75\% missing patches and larger receptive field patches with 25\% visible patches. DEMAE \cite{Li2024demae} devised a diffusion-enhanced MAE by integrating simultaneous denoising and reconstruction tasks.
		
		Contrastive learning (CL) learns representative features by maximizing the distance between image pairs from different classes and minimizing the discrepancy between different augmented views of the same image. One essential component of CL is data augmentation, which can be categorized into spatial augmentation and spectral augmentation \cite{rs15061612}. Spatial augmentation operations include geometric transformations (random cropping, scaling, rotation, and horizontal/vertical flipping) and color distortions (brightness, contrast, blurring, graying, saturation, color jitter, normalization, and hue). In contrast, spectral augmentation contains random cropping, random block discarding, reverse, spectral rotation, and Gaussian noise addition \cite{Huang2022swinvit, Kang2023ssloil}. MSVT \cite{Chen2021multistageViT} combined the augmented sample with the original ones as the stacked virtual samples for training. Moreover, a dynamic feature augmentation, which randomly masked features and the coordinates of the mask dynamically changed during training, was proposed by \cite{rs13030498}. 
		
		Various works have been devoted to HSI classification tasks with Transformers using contrastive learning schema. For instance, 3DSwinT \cite{Huang2022swinvit} performed contrastive learning on the multiscale outputs of the four-stage 3D Swin Transformer blocks, which composed the Siamese structure of the network. MSMT-LCL \cite{Zhou2024msmt-lcl} adopted the multiscale augmented feature mapping to project original HSI data into two feature maps, which are then separately processed by two masked Transformer branches for reconstruction. SSACT \cite{rs15061612} divided the hyperspectral data into the visible light range and the non-visible light range. Then, the first three bands with the largest information entropy values in the two ranges are obtained as the input of spatial contrastive learning. At the same time, the spectral curves of the target pixels were also augmented for spectral contrastive learning. 
		
		CViT \cite{Zhou2023ViTContrast} constructed unsupervised contrastive learning by augmenting different views of the same sample. It utilized supervised contrastive learning by exploiting features of different samples from both the same class and different classes. To address the domain shift issue, EHSnet \cite{Wang2024EHSnet} developed a domain generalization (DG) method to encode image and text features in the semantic space for supervised contrastive learning. NeiCoT \cite{LIANG2024103979} utilized contrastive learning for tokenization and proposed a sequential-specific masking scheme to prompt the Transformer encoder to perceive latent features. Moreover, although CL and MAE were usually introduced separately for HSI classification, they may extract complementary information from the data. Therefore, TMAC \cite{Cao2023maskedAECL} proposed to fuse the features extracted by an MAE branch and a CL branch.

		\item Other learning schemas are not so commonly adopted by HSI classification tasks with the Transformer to handle the limited sample problem. For instance, \textbf{Active learning} selects the most informative samples for training under specific criteria to save the labor for expert annotating. Collaborative AL (CAL) \cite{Wang2023improvedCapsule} combined Capsule Network with ViT in an active learning scheme, while MAT-ASSAL \cite{Zhao2023multiattention} iteratively selected the most informative samples by a sample query rule and derived pseudo labels with high confidence from the corresponding segmented superpixels. 
		
		In addition, ITER \cite{Yang2023iter}, PUSL \cite{Yao2023pusl}, and SCM-CT \cite{Zhao2024SCM-CT} generated pseudo labels for \textbf{semi-supervised / weakly supervised learning} HSI classification, where semi-supervised learning used a mixture of labeled and unlabelled data, while weakly supervised learning used partially or imprecisely labeled data. ITER was designed to achieve image-level HSI prediction, PUSL aimed to explore the rich information in the low-confidence unlabeled samples, and SCM-CT, which contained one supervised learning and one unsupervised learning module, proposed a self-adaptive threshold and conflict pseudo-labeling (SATCP) strategy to ensure that the two learners generate more confident pseudo-labels. 
		
		CTF-SSCL \cite{Xi2024CTF-SSCL} integrated semi-supervised learning and contrastive learning by combining supervised and unsupervised contrastive learning, whose sample pairs were generated by random perturbation of the spectral features. SPTNet \cite{Ma2024SPTNet} designed the noise-tolerant learning algorithm by dividing the training data into clean and noisy sets. This design aims to analyze the resemblance between clean and noisy samples using similarity regularization.
		
		Moreover, HyperViTGAN \cite{He2022HyperViTGan} was designed to address the imbalance issue of HSI data by incorporating a \textbf{generative adversarial network} with Transformer and an external semi-supervised classifier. The real spectral properties were superimposed to generate the augmented data during the discriminator training. Without external datasets, additional noisy HSI data that was output by an external generator was used with the original data to enhance classification performance \cite{rs14143426}. In TRUG \cite{Hao2023trug}, a residual upscale method was proposed to increase the resolution of generated images gradually.

	\end{enumerate}
	
	The techniques mentioned above to address the limited sample issue have unique advantages and disadvantages. Although utilizing cross-domain information can improve classification performance according to the reviewed works' experiments, the effectiveness of this approach is limited by the quality and quantity of labeled samples and the significant spectral shift between the target domain and source domain \cite{Yuan2021pretrainvit, rs14143426}. Adopting models pre-trained on large-scale RGB datasets, such as ImageNet, results in additional work to reduce spectral channels, which can cause significant information loss of HSI data, and the pre-trained model size can be unnecessarily large for HSI tasks \cite{rs13030498, Wang2023DCN-T}. In addition, the visual information of RGB source domains might be too distinct from the HSI target domains \cite{Zhao2022crossscene}. Thus, the performance might deteriorate, or additional computational power must be devoted to domain alignment. Unlike large-scale RGB datasets that contain hundreds or even thousands of different classes, HSI datasets, such as Chikusei \cite{NYokoya2016} or Indian pines, do not contain sufficiently diversified target categories. 
	
	Moreover, the issue of lacking appropriate source domain datasets is more severe for applications on proximal HSI classification tasks, including food quality control and medical image analysis. For instance, considering the variety of diseases, it is unreasonable to utilize the public datasets of brain tumor tissue \cite{Leon2023brain} and cholangiocarcinoma \cite{Zhang2019Cholangiocarcinoma} as source domain for other medical image diagnosis tasks. Additionally, these datasets were mainly adopted for semi-supervised / weakly-supervised learning tasks \cite{GAO2023107568, ZHANG2024108042}. As for the agriculture areas, researchers usually had to collect specific samples and utilized part of these samples as the source domain and the rest as the target domain, such as soil total nitrogen assessment of different particle sizes \cite{ZHAO2025110409} and cadmium content detection in oilseed rape leaves under different silicon environments \cite{ZHOU2025143799}.
	
	Furthermore, the computational cost for self-supervised learning is expensive \cite{Penarrubia2025}, especially for those that utilize dual-channel architecture to learn the representation from positive and negative sample pairs. The choice of distance metric also affects the performance \cite{Gui2024sslreview}. Due to the selection of highly confident samples, the classifiers of active learning usually suffer from overfitting and lack generalisability. 
	
	The classification performance of semi-supervised learning is sensitive to the label quality, while the label noise negatively affects the model performance under a weakly-supervised learning schema \cite{Yao2023pusl, Zhao2024SCM-CT}. Furthermore, there is no guarantee that the generated samples of GANs can improve the generalisability of the discriminator because the real-world interference might be too diverse to simulate with the generator \cite{PALLOTTINO2025109919}. Therefore, more efforts should be devoted to tackling the challenge of insufficient samples. 
	
	Another area for improvement in HSI classification is the imbalanced samples among different classes. HyperViT \cite{Praven2022HyperViT} and LGGNet \cite{Zhang2023LGGNet} implemented an oversampling method to address the issue of imbalanced samples by duplicating data points from a minority class to roughly match the number of samples from the class with the most samples. HyperSFormer \cite{rs15143491} designed an adaptive min log sampling (AMLS) strategy, which determined the number of samples for class $k$ as: 
	
	\begin{equation}
		Sample_k = (\log_2 (\frac{num_k}{num_{min}})+1)*num_{min}*s
	\end{equation}
	
	\noindent where $num_k$ is the total number of class $k$, $num_{min}$ is the number of class with minimum samples, and $s$ is sampling factor. In order to improve accuracy, a co-learning strategy was adopted, where the CNN and Transformer in the dual-architecture iteratively select highly reliable test samples and combine them with training samples of each other \cite{Chen2022colearning}.

	\section{HSI Preprocessing Techniques}\label{prep}
	
	In order to reduce computational cost and extract shallow features, many HSI classification tasks utilized PCA for preprocessing. \cite{Arshad2024hierarchi} utilized PCA to reduce the spectral dimension while maintaining the spatial dimension to focus on extracting local and global spatial features with hierarchical attention. \cite{Cao2022mixedres} developed a Channel Shift technique on PCA, which moves the important spectral channels to the central position of the data while placing the less important spectral channels at the edge of the data. 
	
	MHCFormer \cite{Shi2023mhcformer} claimed that factor analysis (FA), which models HSI as a combination of shared factors and unique factors, may be more suitable than PCA for dimensionality reduction of HSI data so that the variability in correlated spectral bands can be better captured. Spa-Spe-TR \cite{He2022twobranch} used a linear mapping layer to reduce the input dimension. However, the simple linear projection might lose non-linear information contained in HSI data. 
	
	Other techniques, including normalization and band selection, were also adopted for HSI classification tasks. For instance, \cite{Xu2023ss1dswin} applied channel-wise MinMax normalization and 2-D mirror padding to preserve continuity at the edges of the image. However, MinMax normalization might not be appropriate for the HSI classification task \cite{Zhang2023OptimizingHI}. In addition, SPFromer \cite{Li2023spformer} adopted a two-layer autoencoder, and LAGAN \cite{CHEN2023120828} used a 1D-CNN-based autoencoder for dimensionality reduction with mean square error loss (MSE). Although these autoencoder-based dimension reduction methods can capture non-linear features well, they add extra computational complexity for further processing. Moreover, these feature extraction-based dimension reduction methods, including PCA, linear mapping, and autoencoder, corrupt the spectrum property, thus making the extracted features physically meaningless \cite{ZHANG2023100491}. 
	
	In contrast, other authors conducted band selection to reduce the HSI dimension. \cite{Tulapurkar2022mha} implemented a band ranking method based on the classification accuracy of each training epoch, while \cite{Tejasree2023} used a grey wolf optimizer (GWO) for band selection based on the features extracted by t-distributed stochastic neighboring embedding (t-SNE). A tensor-based Sobel edge detection method was implemented to avoid the noise of HSI and regularize the contour of the image \cite{Hao2023multilayer}. A band selection method based on the differences between the signal-to-noise ratio (SNR) of the two spectra at each wavelength was developed for glaucoma detection \cite{diagnostics14121285}. The Crow Search Optimization (CSO) technique was adopted with Bhattacharya distance calculation by DNAT \cite{Tejasree2024dnat} for band selection. DISGT \cite{Cheng2024disgt} designed a dynamic distance norm to measure the correlation between bands for unsupervised band selection. The authors of EggFormer \cite{JI2024109298} compared the performance of different band selection techniques, including Random Forest (RF), PCA, Successive Projections Algorithm (SPA), and Competitive Adaptive Reweighted Sampling Algorithm (CARS).

	\section{Sample generation and embedding} \label{embed}
	
	Due to the high dimensionality, HSI data was usually split into small patches and converted into token embeddings to be processed by Transformer-based networks. In addition, positional embedding is often added to enhance the spatial information for HSI feature extraction. This section will discuss different patch-splitting methods, token embedding generation approaches, and various positional embedding types. 
	
	\subsection{Patch splitting methods}
	
	There are mainly three sampling strategies: pixel-wise, which uses a group or all bands of a pixel as input; patch-based, which adopts a fixed-size window around the targeted pixel as input; and image-based, which uses the whole image with a given number of labels as input \cite{YANG2023145}.

	\begin{enumerate}
		\item \textbf{Pixel-wise}: Few studies adopt pixel-wise HSI classification. CSiT \cite{He2022CSiT} was designed to learn multiscale features on spectral sequences, MCE-ST \cite{KHOTIMAH2023103286} took spectral signals as input for crop stress detection, and ViT was utilized on pixel-level brain-tissue HSI images because HSI patches did not provide a significant improvement of the results according to experiments \cite{Cruz-Guerrero2023braintissues}. The pixel-wise input was rarely adopted because the spatial information also contributed significantly to the classification performance \cite{Liu2022central}.
		
		\item \textbf{Patch-wise}: Most works used HSI patches as input because the spatial information can be as influential as spectral information in HSI classification \cite{Liu2022central, Xu2020beyond}. The common practice is to generate a fixed-sized window around the target pixel and pad the edge pixels \cite{Zhao2022contrans, Sun2022tokenization, Qian2023neighborhood, Zhang2023ELS2T}. For a target pixel (the pixel for which we want to predict the label), a region (context) that contains the target pixel is selected and flattened into a pixel sequence. In order to be exposed to many different contexts (regions) of the target pixel for better generalizability, HSI-BERT \cite{He2020hsibert} can take arbitrary regions as input and extract features by flattening these regions into a pixel sequence and then padding to a maximum length with dummy pixels. 
		
		To fully use the spatial neighborhood information of the targeted pixels and reduce the computational complexity, the following patch method was proposed by calibrating the patch coordinates without using padding \cite{Yang2023gtfn}. In order to determine the most appropriate window size for splitting HSI cubes, BS2T \cite{Song2022bs2t} compared the effects of input patch size on various datasets. 
		
		MATA \cite{Liu2023multiarea} and SSTE-Former \cite{Wu2023tokenhash} implemented multiple HSI patches of different sizes (multiscale) centered on the target pixel. MATA \cite{Liu2023multiarea} also divided these patches of different sizes into areas of different directions (up, down, left, and right) to identify objects in different areas. A hierarchical region sampling strategy \cite{YANG2023145} was proposed to simultaneously generate the center region, neighbor region, and the surrounding region, which separate Transformers processed to extract features of different granularities. LSDnet \cite{Peng2024LSDnet} proposed a bilateral filtering-based feature enhancement (BFFE) module to enhance the spatial features consistent with the center pixels and suppress the pixels' features inconsistent with the target pixels.
		
		Since different arrangements of inputs in a sequence can lead to varied output features for sequential models, RNN-Transformer \cite{Zhao2023rnntrans} utilized the multiscanning strategy, which scanned the HSI sequence in multiple different orders, to incorporate the RNN's ordering bias with the Transformer's global feature extraction. MaskedSST \cite{Linus2023marsked} and spatial-spectral-based 3D ViT \cite{Zhou2022rethinking} divided the input image into spatial-spectral patches, which split HSI cubes into spatial and spectral dimensions simultaneously.  
		
		However, the patch splitting and padding operations usually include interference from irrelevant information around the target pixel, especially in the boundary area. Moreover, the overlapping patching may lead to information leakage that the training patches may contain neighboring or target pixels from test patches \cite{Bai2022multibranch, Zhang2023MATNeT}. Because large patch-splitting windows include too much neighboring information, leading to non-independent training and testing sets, \cite{rs15163960} proposed to randomly shuffle the neighboring pixels around the central pixel within a small window to obtain more independent training and testing sets. 
		
		Nevertheless, the pixels in the test set might still be included in the training set for this method. Therefore, researchers proposed various methods to improve the patch-splitting procedure. In order to reduce the overlap between training and test samples, LiT \cite{Zhang2023lightweight} developed a controlled multiclass stratified (CMS) sampling strategy. The training and testing sets were determined according to whether all pixels had the same labeled class within each HSI patch and iteratively masked out classified pixels. \cite{Viel2023analysis} proposed a row and column division process, which allocated even rows and columns for training and odd rows and columns for testing. Nonetheless, this method decreased the number of samples that could be used for training.  
		
		\item \textbf{Pixel-wise and patch-wise}: Some works utilized both single pixel and pixel patches as input by extracting spatial features from image patches and spectral features from pixels with a two-branch architecture network \cite{Xin2022conhsi, Lu2023multiattention, Kumar2023methanemapper}. For instance, DSS-TRM \cite{Liu2022dsstrm} used a convolution layer to transform the image blocks of different bands into a one-dimensional feature vector for the spectral dimension. It split the entire HSI image into 16 patches with the same size spatially as in \cite{dosovitskiy2021an}, and mapped the 16 patches into 16 one-dimensional feature vector with a convolutional layer. 
		
		Spa-Spe-TR \cite{He2022twobranch} also split the HSI sample into several non-overlapping patches after linear projection to reduce the dimension for the spatial Transformer branch. The pixel vector is fed to the spectral Transformer for modeling long-range dependencies. The GBiLSTM network \cite{Xu2022biLSTM} used a single pixel vector for spectral feature extraction in a grouped manner. S2FTNet \cite{Liao2023fusion} fed image patches to the Spatial Transformer module for spatial feature extraction after reducing the spectral dimension by PCA and pixels to the Spectral Transformer for spectral feature extraction. These methods decoupled the relationship between spatial and spectral features. Thus, additional feature fusion modules became crucial for the classification performance.
		
		\item \textbf{Image based}: Traditional patch-based framework, which suffers from limitations such as restricted receptive fields and high computational complexity \cite{Li2023generative}, is replaced with the image-based classification framework to solve the inefficiency in the training and testing \cite{Yu2022mstnet} with faster inference speed \cite{Shi2023g2t}. A binary mask of the same size as the HSI image was used to select pixels for training, then the trained model outputs the predicted labels of all pixels in the whole image \cite{Yu2022mstnet, Yang2023multilevelIT, Yan2023hybridconvvit}. According to the experimental results, the image-based classification framework can significantly reduce the training and inference time because the redundancy from overlapping patches was not included in this learning framework. Moreover, the selection of optimal patch size is unnecessary. However, the spatial information from neighboring pixels that can assist inference during testing may still leak to the training sets.
		
		\item \textbf{Clustering}: Another approach to segment HSI is superpixel, which aggregates similar pixels into irregular subpatches using a space clustering algorithm, such as Simple Linear Iterative Clustering (SLIC) \cite{Zou2022LessFormer, Feng2022gaussian, CHEN2023120828, Feng2024cat, Huang2024APSFFT}. Spatial Sample Selection(3S) mechanism \cite{Feng2022gaussian} was proposed to reduce the interference of heterogeneous pixels by selecting neighboring pixels only from homogeneous regions. G2T \cite{Shi2023g2t} utilizes linear discriminant analysis (LDA) along with SLIC to construct a superpixel graph. However, different objects might present the same spectrum in the HSI dataset, while the same target objects might demonstrate a distinct spectrum, thus decreasing the quality of the superpixel segmentation. 
		
	\end{enumerate}

	\subsection{Sequence Generation Using Token embedding}
	
	The pixel or patches from HSI should be converted into a sequence to be compatible with the input of the Transformer block for the following feature extraction \cite{Sigirci2022hyperslic}. Usually, a class token and positional embedding were also added to the token embedding so that image sequence embeddings can be processed as token (word) embeddings in NLP applications \cite{dosovitskiy2021an, Zu2023cascaded}. 
	
	Most of the research in this field embedded the inputs by unfolding the 2-D patches and using \textbf{linear} projections to embed vectors \cite{Liu2022coastal, He2022twobranch, Qiao2023sstf, Zu2023wrsag, Xue2022partition}. Because the adjacent spectral bands are correlated, SpectralFormer \cite{Hong2022spectralformer}, HSST \cite{Song2022hierarchical}, S3FFT \cite{Xie2023fusion}, and SSPT \cite{Ma2023tokenreduction} proposed learning groups of several adjacent channels for spectral embeddings and converting the grouped tokens to embedding vectors by a trainable linear projection. 
	
	In addition, the spectral channels can be aggregated into different group numbers to generate multiscale embeddings. CSiT \cite{He2022CSiT} transformed all bands in the spectral dimension with linear projections into multiscale representations on different numbers of band groups. MSTViT \cite{Li2022granu} divided the feature map into different-granularity tokens and embedded these tokens in parallel by a linear projection. As dipicted in Fig. \ref{morphformer}, morphFormer \cite{Roy2023morpho} used two learnable weights to generate two independent projected patches, then multiplied the projected vectors as a tokenization operation. 
	
	The linear embeddings can also be performed on both the spectral and spatial dimensions. DATE \cite{Tang2023double} obtained spectral tokenization by unfolding input patches along the spatial direction and applying a learnable linear transformation. It acquired spatial tokenization by splitting spatial information of each band into non-overlapping submatrices and applying another learnable linear transformation to the flattened submatrices. Moreover, as shown in Fig. \ref{ssftt}, SSFTT \cite{Sun2022tokenization} and MATNet \cite{Zhang2023MATNeT} tokenized the extracted features by conducting a 1$\times$1 pointwise product with a weight matrix initialized with a Gaussian distribution. The linear projection is simple and broadly applied, whereas it can lose the non-linear structural information in HSI data. 
	
	\textbf{Convolutional} layers can generate richer tokens, introduce some inductive bias properties of CNN into the Transformer \cite{WANG2022103005, Zu2023cascaded}, and preserve more local information. In contrast, linear projections cannot capture the structural information present in patches \cite{Zhang2023lightweight}. Thus, a patch embedding module that utilizes convolutional operations is commonly utilized to divide the input images into multiple patches and convert them into 1D tokens \cite{Zhou2023dictionary}. 
	
	For example, GAHT \cite{Mei2022groupaware} fed HSI patches into grouped pixel embedding, which used 1$\times$1 convolutions to split patches into pixel sequences as input for the Transformer block. CESSUT \cite{Xin2022conhsi} utilized a 2D convolutional layer on spatial patches for spatial embedding and 1D convolutional on spectral vector for spectral embedding. MCE-ST \cite{KHOTIMAH2023103286} implemented a 1D convolutional layer to learn local features from the raw spectral information, followed by an average-pooling layer to reduce redundant information and a linear layer to enrich each token representation. LESSFormer \cite{Zou2022LessFormer} generated token embedding using a shallow CNN mainly consisting of two 1$\times$1 convolutional layers, a group convolutional block containing the 1$\times$1 group convolutional layer, and a learnable superpixel segmentation submodule of SLIC for dividing patches into irregular subpatches. 
	
	In order to embed patches of different scales, DFTN \cite{Qiao2023fdfe} employed a 1$\times$1 2D convolutional layer to embed multiscale patches. LSDNet \cite{Peng2024LSDnet} used four convolutions of different sizes to output multiscale spectral embeddings. HSD2Former \cite{rs16234411} adopted multiple 3D convolutional layers of different kernel sizes to generate multiscale spatial and multiscale spectral embeddings. 
	
	Moreover, \textbf{3D convolution}, which moves in three directions, is widely utilized for 3D HSI patch embedding and shallow feature extraction \cite{Li2023cnntrinter}. For instance, Spectral-MSA \cite{Zhou2021swinspectral} adopted a 4$\times$4$\times$1 3D convolutional layer for patch embedding. MAR-LWFormer \cite{Fang2023multiattention} utilized different 3D convolution kernels to output multiscale tokens. HybridFormer \cite{Ouyang2023hybrid} performed patch embedding through depthwise convolution, which can partition the feature map into several semantic tokens with different granularities to keep the spatial and spectral information independent. GTCT \cite{Qi2023globallocal} generated spectral embedding and spatial embedding in a parallel way using different 3D convolutions spatially or along spectral dimensions. 
	
	Many works also utilize combinations of 2D and 3D convolutions to embed HSI patches. For example, LSGA-VIT \cite{Ma2023gaussian} first utilizes the 3D convolutional layer to extract spatial-spectral features, whose first two dimensions were merged and fed to 2D convolution. MCTT \cite{Wang2024MCTT} extracted spatial-spectral features with a sequence of 3D, 1D, and 2D convolutions. Then the flattened features $F$ was tokenized to $\hat {T}$ by multiplying with two learnable matrices $W_{1}$ and $W_{2}$ as $\hat {T}=\left ({F \cdot W_{1} }\right) \cdot \text {softmax} \left ({F \cdot W_{2} }\right)^{T}$. After PCA for dimension reduction of HSI cubes, LSFAT \cite{Tu2022aggregation} utilized the 3D convolutions to extract feature cubes, which were flattened and fed to a linear projection to generate embedded tokens. IMAE \cite{Kong2023instoken} sequentially implemented a 1$\times$1 2D convolutional layer, a 1$\times$1$\times$1 3D convolutional layer, and another 1$\times$1 2D convolutional layer to unify spectral dimension and explore spectral information. WaveFormer \cite{Ahmad2024waveformer} developed wavelet transformation as downsampling to decompress 3D feature maps that 3DCNN processes. 
	
	Although the 3D convolutional layer is straightforward to process the three-dimensional HSI data cubes and flexible to move along different dimensions, it can significantly increase the model parameters \cite{Hong2022spectralformer, Zhang2023ELS2T}, leading to high computational costs and overfitting \cite{Xiao2023dsdcdcam}.

	\subsection{Positional embedding}
	
	Since the Transformer-based models usually employ 1D sequences as input, the positional relationships of the sequences are too complex to capture \cite{Vaswani2017attention}. Without the function of recording position information, the self-attention mechanism can easily cause disorder during feature fusion \cite{Cui2023DAFFN}. Therefore, proper positional encoding (PE) is crucial for HSI classification tasks, as positional embedding significantly impacts the Transformer’s structural comprehension and the provision of ordering information \cite{Liu2020icml}. The PE vector was usually added or concatenated to the feature embedding or added as a bias to the self-attention \cite{Chen2023projection}.  
	
	The original work that introduced the self-attention mechanism implemented sine and cosine functions of different frequencies as the positional embedding \cite{Vaswani2017attention, Wu2021reposemb}:
	
	\begin{equation}
		\begin{array}{l} 
			{{\text{P}}{{\text{E}}_{pos,2i}}\quad \;\; = \sin \left( {pos/{{10000}^{2i/d}}} \right)} \\ {{\text{P}}{{\text{E}}_{pos,2i + 1}}\quad = \cos \left( {pos/{{10000}^{2i/d}}} \right)} 
		\end{array}
	\end{equation}
	
	\noindent where $pos$ is the position, $i$ represents the current dimension of the positional encoding, and $d$ is the feature dimension. Many works added this positional embedding to each token \cite{Liu2022dsstrm, Zhang2023ELS2T, Kumar2023methanemapper}. \cite{Feng2022gaussian} proposed Gaussian Positional Embendding(GPE) by multiplying the sine and cosine functions with a Gaussian weight as: 
	
	\begin{gather}
		\text{GPE}_{(m,2n)}=\sin \left(i/10000^{2n/d} \right)\times G_{n}\\
		\text{GPE}_{(m,2n+1)}=\cos \left(i/10000^{2n/d} \right)\times G_{n}\\
		\text{G}_{n}=e^{-((x-x_{0})^{2}+(y-y_{0})^{2})/(2\sigma^{2})}
	\end{gather}
	
	\noindent where $m$ is the band number and $n$ is the pixel number. $G_n$ is the $n$th gaussian weight, and $(x_{0}, y_{0})$ is the center pixel coordinate. This positional embedding encodes pairwise relationships between elements \cite{Wu2021reposemb}, making them translation invariant. In addition, some works incorporated the position information by adding a relative position bias matrix $B$ to each head in MHSA, as 
	
	\begin{equation}
		\textbf {Attention}\left ({Q,K,V}\right)=\textbf {Softmax}\left (QK^{T}/\sqrt{d}+B \right )V
	\end{equation}
	
	\noindent The relative position bias is generated according to the relative distance between pixel tokens in subpatch \cite{Zhou2021swinspectral, Selen2022SpectralSWIN, Xue2022partition, Qian2023neighborhood, Qiao2023fdfe}. AttentionHSI \cite{rs14091968} included the relative position bias for spatial attention inside the self-attention as 
	
	\begin{equation}
		\textbf {Attention}_{spatial}\left ({Q,K,V}\right) = \textbf{Softmax} \left( QK^T/\sqrt{d} \right) + BV
	\end{equation}
	
	\noindent LSGA-VIT \cite{Ma2023gaussian} replaced the positional bias matrix $B$ with a Gaussian location matrix, which used a 2D Gaussian function to represent the HSI spatial relationship. In Spa-Spe-TR \cite{He2022twobranch}, the positional embedding in the spectral Transformer branch was generated from a normal distribution and added to the extracted features. BS2T \cite{Song2022bs2t} utilized relative positional encoding with spectral information as:
	
	\begin{equation}
		\textbf {Attention}\left ({Q,K,V,S}\right) = \textbf {Softmax}\left ( ({R_{h}+R_{w}} )QK^{T}/\sqrt {d} \right)SV
	\end{equation}
	
	\noindent where $S$ is the spectral feature obtained by global pooling the entire spatial feature on a channel, while $R_h$ and $R_w$ are the relative position encodings for height and width, respectively. The relative positional encoding was also implemented in \cite{Zhang2023D2S2BoT} as: 
	
	\begin{equation}
		\textbf {Attention}\left ({Q,K,V}\right)=\textbf {Softmax}\left ( Q({R_{h}+R_{w}} ) + QK^{T}/\sqrt {d} \right)V
	\end{equation}
	
	\noindent A 3D coordinate positional embedding \cite{Zhou2022rethinking} was implemented to encode the positional information of feature tokens. 
	
	GrphaGST \cite{Jiang2024GraphGST} established an absolute PE (APE) to generate absolute positional sequences (APSs) for pixels. Given an image of height $h$ and width $w$, which was divided into patches of size $a\times a$, the absolute sequence number $\mathbb{S}_n$ for these patches was calculated as:
	
	\begin{equation} 
		\mathbb{S}_n = {\mathrm {Ceil} \left ({\frac {h}{a}}\right) \times \mathrm {Ceil} \left ({\frac {w}{a}}\right)} \quad \forall \mathbb{S}_n \ll h \times w 
	\end{equation} 
	
	Moreover, ViT \cite{dosovitskiy2021an} introduced learnable 1D positional embedding to model 2D spatial information for input elements, and the learnable positional embedding was adopted by many works in this area \cite{Hao2022featurefusion}. Most works added the learnable embedding to the sequence vectors in an element-wise fashion \cite{Yu2022mstnet, Zhang2022mixer, Roy2023morpho, Qiao2023sstf, Xu2023ss1dswin}, while \cite{Bai2022multibranch} proposed to multiply position weight with all of the patch vectors. 
	
	\cite{Linus2023marsked} compared the learnable and absolute spectral positional embeddings with fixed sine and cosine functions. LGGNet \cite{Zhang2023LGGNet} introduced Gaussian-initialized learnable positional prompting. RNN–Transformer \cite{Zhao2023rnntrans} generated multiscanning-controlled positional embedding through the U-Turn scanning pattern because adjusting the positional information for the Transformer can yield different output features. IMAE \cite{Kong2023instoken} proposed Conditional position embedding, which was implemented by a 2D convolution layer and the same padding layers on the output of spectral embedding. 
	
	Additionally, CTN \cite{Zhao2022contrans} adopted center position encoding by multiplying a parameter matrix with a position matrix, which was obtained according to the relative position of neighboring pixels to the center pixel within each HSI patch. \cite{Wu2023tokenhash} proposed a hash-based positional embedding through the hash-based matrix, which was shared among token embeddings with the same position of different scales. SQSFormer \cite{Chen2024SQSFormer} adopted a rotation-invariant position embedding module, which randomly rotates the input patches around the center pixel and uses relative position indices corresponding to the center pixel as Center Relative PE, to alleviate the spatial noise. 
	
	Some research studies also applied \textbf{no positional embedding} \cite{Liu2023multiarea}. For instance, LESSFormer \cite{Zou2022LessFormer} did not adopt positional embedding since features were extracted from unordered superpixels. Due to the high complexity of the spatial distribution of HSI, PUSL \cite{Yao2023pusl} did not encode the spatial position information into spatial-spectral features to avoid overfitting. 
	
	Moreover, some works adopted convolutional operations to preserve and encode the position information \cite{Dang2023DoublebranchFF}, such as depthwise convolution \cite{Wang2023DCN-T, WANG2022103005}. The depthwise convolution performed computation along the channels dimension, which is both parameter and computation efficient \cite{rs14153705}. DAFFN \cite{Cui2023DAFFN} proposed a dual-branch position self-calibration (PSC) module, where one branch kept the original feature information, and the other branch was designed to calibrate the position information with an arbitrarily sized pooling kernel and convolution transformation. Thus, the positional information can be preserved during the convolutional operation, and the two branches are fused with trainable weights. Although these works stated that the convolutional operation can preserve the positional information so that the extra position vector is unnecessary, the convolutional layers cause more computational complexity than a simple learnable vector.

	\section{HSI Feature extraction}\label{featureextract}
	
	Feature extraction is the most crucial component for HSI classification tasks. Researchers in this field proposed a variety of modules to learn the spatial-spectral features from HSI data. These different methods aimed to address various challenges for HSI classification, including the intrinsic correlation between spatial and spectral domain \cite{Zhou2022rethinking}, weak intraclass spectral consistency \cite{Chen2023projection}, high relevance between adjacent spectral bands and low dependencies across long-range ones \cite{Lu2023multiattention}, geometric constraints caused by input data \cite{YANG2023145}, and contributions of features in the decision process \cite{rs14030716}. Due to the diversity and complexity of these techniques, we categorized these feature extraction methods into three main classes: extracting spectral features, extracting spatial-spectral features together within one module, and extracting spatial-spectral features separately with different modules. 
	
	\subsection{Spectral feature extraction}
	
	Although HSI data contains spectral and spatial information, some works only utilized Transformers to learn the relationship along spectral dimensions. For example, a work directly utilized ViT to extract spectral features from pixel-level brain-tissue HSI images because patch-wise data did not provide a significant improvement according to experiments \cite{Cruz-Guerrero2023braintissues}. As shown in Fig. \ref{spectralformer}, SpectralFormer \cite{Hong2022spectralformer} proposed implementing an overlapping grouping operation to learn spectrally local sequence information from neighboring bands of HSI for group-wise spectral embeddings. In addition, after the spatial sample selection block generating image cubes with SLIC, \cite{Feng2022gaussian} designed the spectral feature extraction block to capture long-range information between spectral groups. 
	
	Furthermore, some works have been designed to learn multiscale spectral features. CSiT \cite{He2022CSiT} utilized a small branch to learn small-scale spectral features with fewer encoders and smaller embedding dimensions. It used a large branch to extract local features in coarse-grained with more encoders and larger embedding dimensions. MCE-ST \cite{KHOTIMAH2023103286} extracted local features from the spectral input with 1D convolution and average-pooling operations for crop stress detection. These spectral features were fed to the MHSA module and the Feed-Forward module. Then, the outputs were processed by a point-wise convolution, two depth-wise dilated convolution layers with different kernel sizes (3 and 5) in parallel, and a swish activation function \cite{Ramachandran2018SearchingFA} to learn different spans of interactions between tokens. 
	
	Moreover, SSPT \cite{Ma2023tokenreduction} passed the group-wise linear projection of spectral bands into a series of ViT-based encoders with the same components, which had additional token reduction and compensation modules. GSPFormer \cite{Chen2023projection} mapped all spectra within a patch to a new representation space by the same transformation and then aggregated them at spatial dimensions to enhance the spectral feature of the center pixel. SVAFormer \cite{Chen2024svaformer} employed random masking techniques to enhance the spatial neighborhood information for the spectral tokens from different perspectives. 
	
	Nowadays, most research in this field emphasizes both spectral feature extraction and spatial feature extraction. However, it might be possible that spatial information cannot enhance the classification performance for some datasets \cite{Cruz-Guerrero2023braintissues}. Thus, focusing only on spectral feature extraction might still be worth consideration for researchers in some specific areas.

	\subsection{Extracting spatial-spectral features together}
	
	There are a variety of operations to extract spatial-spectral features altogether. After embedding the HSI pixels or patches, many works directly used the token embeddings for the subsequent processing with Transformer blocks without additional feature extraction, such as HSI-BERT \cite{He2020hsibert}, MSVT \cite{Chen2021multistageViT}, MFSwin-Transformer \cite{agronomy12081843}, SPRLT-Net \cite{Xue2022partition}, and PUSL \cite{Yao2023pusl}. 
	
	Some works aimed to learn multiscale or multi-stage spatial-spectral features with Transformer blocks of different functions. HSST \cite{Song2022hierarchical} passed the embedded patches into three Transformer blocks with similar architectures to learn spatial-spectral features in light, middle, and deep levels by transforming the feature map into cubes with different channels. In addition, after summing the group-wise feature token embedding with a learnable positional embedding, four stages of 1DSwin Transformer blocks \cite{Xu2023ss1dswin}, which concatenated odd and even elements of the input feature map alternatively and linearly mapped to reduce feature tokens, were implemented to extract spatial-spectral features. CSIL \cite{YANG2023145} sent the center pixel's embedding into the center Transformer and combined the neighbor and surrounding region to form the surrounding embedding for the surrounding Transformer. After achieving fine-grained pixel-to-region assignment of superpixels, LAGAN \cite{CHEN2023120828} collected local information and global correlation with modified MHSA modules hierarchically. DiCT \cite{Zhou2023dictionary} fed the embedding into the group self-attention mechanism, which can improve the stability of the weight matrix by grouping neighboring pixels. 
	
	However, the MHSA mechanism can capture the long-range relationships between token embeddings well but might not be able to extract the spatial-spectral features thoroughly. Therefore, utilizing extra feature extraction modules became a straightforward option.  
	
	Numerous works used external feature extraction modules besides Transformer blocks. For example, SEDT \cite{Wu2022densely} only used \textbf{Pooling} operations for feature extraction. After dividing the HSI into patches according to the raster scan order, SEDT conducted global averaging and max-pooling operations on the spatial and spectral information of the neighboring pixels of the central pixel. The summation of the two pooling layers was added to the original patch to enhance the spectral characteristics of the central pixel. After flattening and adding positional encoding, the extracted features were fed to Transformer encoders \cite{Wu2022densely}. Similarly, MASSFormer \cite{Sun2024massformer} implemented memory tokenization on the spatial-spectral features with average-pooling to preserve overall information and max-pooling to emphasize the salient feature points. The output tokens were concatenated with $K$ and $V$ to compute multi-head self-attention.
	
	Besides pooling, other works implemented \textbf{only 2D convolution} to extract spatial-spectral features to be further processed by the Transformer \cite{Zhao2022contrans}. For an image-based classification task, MSTNet \cite{Yu2022mstnet} generated feature maps using a 1$\times$1 and a 5$\times$5 convolutional layer to reduce the spatial size of the original image. SST-M \cite{Bai2022multibranch} used two 1$\times$1 convolution layers for spectral feature extraction before splitting into patch sequences. CDSFT \cite{Qiu2023cdsft} first used three 1$\times$1 convolutions to compress the spectral channels. Then, it adopted 2D depthwise convolution operations with different kernels and channels to extract local spatial features. GAHT \cite{Mei2022groupaware} implemented 1$\times$1 grouped convolutions to divide the channels of feature maps into non-overlapping subchannels and extract spatial-spectral features. DCN-T \cite{Wang2023DCN-T} generated a tri-spectral image, which was fed into a modified VGG-16 backbone network with the ImageNet pre-trained parameters. It also used an extra convolutional layer to produce features for the subsequent Transformer block. MSNAT \cite{Qian2023neighborhood} first adopted 2D convolution to reduce the spectral dimension of HSIs, then used 2D convolutions and max-pooling in spatial transformation (ST) modules to handle spatial variability. Nevertheless, the 2D convolutional operation cannot fully capture the local three-dimensional spatial-spectral features.  
	
\begin{figure}
		\centering
		\includegraphics[scale=0.5]{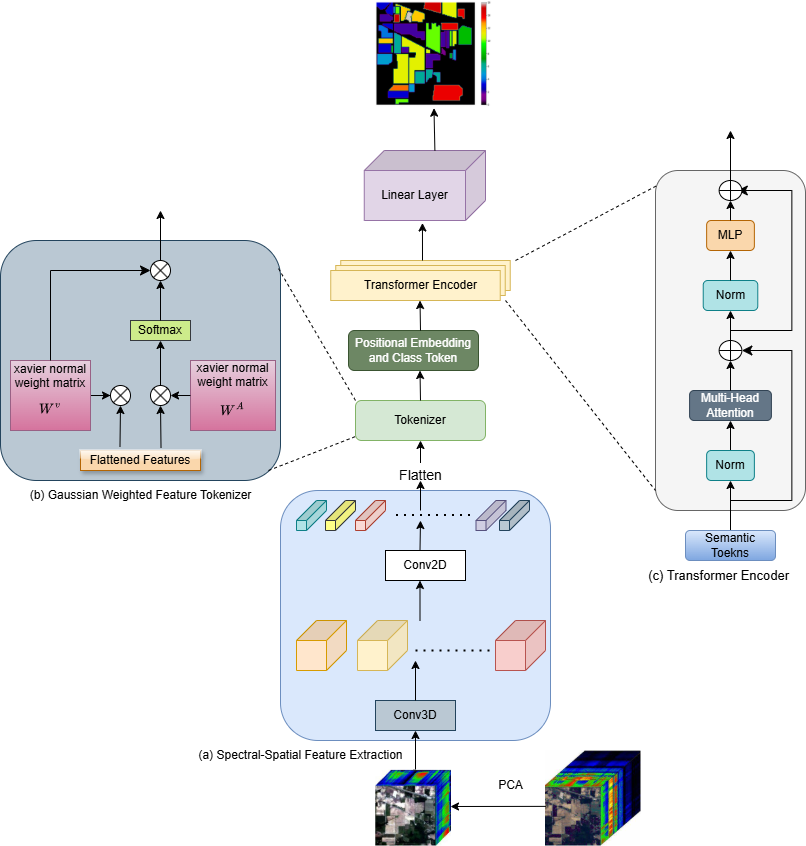}
		\caption{The overall framework for SSFTT \cite{Sun2022tokenization}. PCA is first applied to reduce the spectral dimension of the HSI hypercube. After PCA dimension reduction, the HSI data is divided into 3D patches, which are processed by (a) subsequent 3D convolution and 2D convolution for spectral-spatial feature extraction. The extracted features are then flattened and tokenized by (b) Gaussian Weighted Feature Tokenizer, where the features are multiplied with weight matrices initialized by Gaussian distribution. The generated semantic tokens serve as inputs to the (c) Transformer encoder. The outputs of the Transformer encoder are linearly projected to generate the classification prediction. }  
		\label{ssftt}
	\end{figure}

	Therefore, \textbf{3D convolution}, which can move along three directions of the HSI cubes, was also commonly adopted for spatial-spectral feature extraction \cite{Zhou2021swinspectral}. In order to generate feature maps for subsequent processing with MHSA, some works mainly utilized 3D convolution for feature extraction \cite{s22103902, Tu2022aggregation, Wang2024MCTT}, while others mixed 3D and 2D convolutions to extract spatial-spectral features \cite{Cao2022mixedres, Sun2022tokenization, Zu2023cascaded, Ma2023gaussian, Zhou2023mdvt}, as shown in Fig. \ref{ssftt}, which illustrates the framework of SSFTT \cite{Sun2022tokenization} as an example. Moreover, SDFE \cite{rs15010261} extracted shallow features via two 3D convolutional layers and then used depth-wise convolution and channel attention to strengthen the critical channel information. In order to suppress the large and small feature values and enhance the middle feature values, CITNet \cite{Liao2023integrated} fed the spatial-spectral features extracted by 3D and 2D convolutions through Channel Gaussian modulation attention module (CGMAM), which modulated the distribution of features with Gaussian function. However, the module parameters and computational costs are higher for 3D convolution operations than for 2D convolutions. Additionally, it is worth exploring whether enhancing the spatial and spectral feature learning separately can improve classification performance. 
	
	\subsection{Separately extracting features}
\begin{figure}
		\begin{subfigure}{0.23\textwidth}
			\centering
			\includegraphics[scale=0.3]{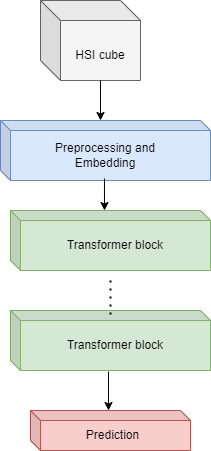}
			\caption{}
			\label{casc}
		\end{subfigure}
		\hfill
		\begin{subfigure}{0.23\textwidth}
			\centering
			\includegraphics[scale=0.3]{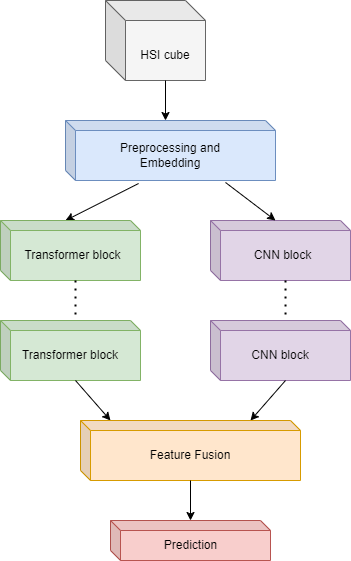}
			\caption{}
			\label{pct}
		\end{subfigure}
		\hfill
		\begin{subfigure}{0.23\textwidth}
			\centering
			\includegraphics[scale=0.3]{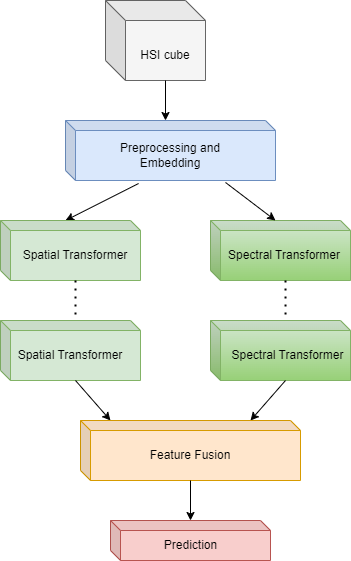}
			\caption{}
			\label{ptt}
		\end{subfigure}
		\hfill
		\begin{subfigure}{0.23\textwidth}
			\centering
			\includegraphics[scale=0.3]{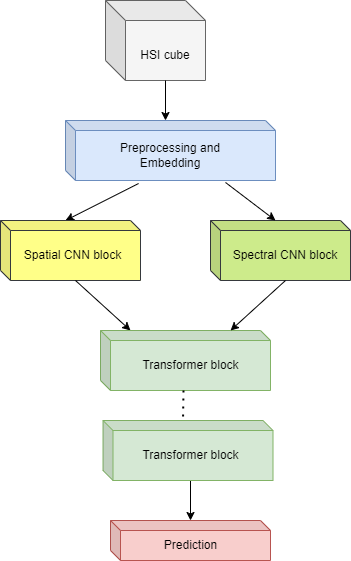}
			\caption{}
			\label{pcct}
		\end{subfigure}
	\caption{An illustration of different feature extraction architectures: (a) cascaded Transformer blocks; (b) parallel CNN and Transformer branch; (c) parallel Transformer branches; (d) parallel modules and subsequent cascaded Transformer. In practice, CNN and other modules could be added within the Transformer branch. Multiple branches could be designed to extract other features besides spatial and spectral features. }
	\label{archi}
	\end{figure}

	In order to better extract the spatial-spectral features from HSI data and efficiently learn the local-global relationships, abundant research in this area has developed various models to accomplish the HSI classification task with separate modules. Different architectures are illustrated in Fig. \ref{archi}.
	
	\begin{enumerate}
		\item Some works extracted spatial-spectral features \textbf{separately in sequential/cascaded architectures}, as shown in Fig \ref{casc}. For instance, SST \cite{rs13030498} used a pre-trained VGGNet to extract the spatial features of each 2D patch. Then, the output was sent to cascaded Transformer encoders to obtain the relationship of the spatial-spectral features. ToMF-B \cite{Zhao2022crossscene} extracted local feature maps with four 2D convolutional blocks with 3$\times$3 kernel size, followed by a Transformer to build the long-range contextual relationship. In order to enhance learning of local spatial relationships, LESSFormer \cite{Zou2022LessFormer} passed the output token embedding into a local Transformer encoder, which emphasized the local interactions between neighboring pixels with a mask matrix, and a standard Transformer encoder to learn global relationships in serial. CMT \cite{Jia2024CMT} proposed a regularized center-masked pretraining (RCPT) to effectively learn the dependencies between the central object and its neighboring objects using center pixel reconstruction and sample reconstruction with Transformer encoders. As displayed in Fig. \ref{datn}, DATN \cite{SHU2024107351} sequentially calculated spectral self-attention on the spectral tokens and spatial self-attention along the spatial patches.
		
		\begin{figure}
			\hspace{-0.6in}
			\includegraphics[scale=0.5]{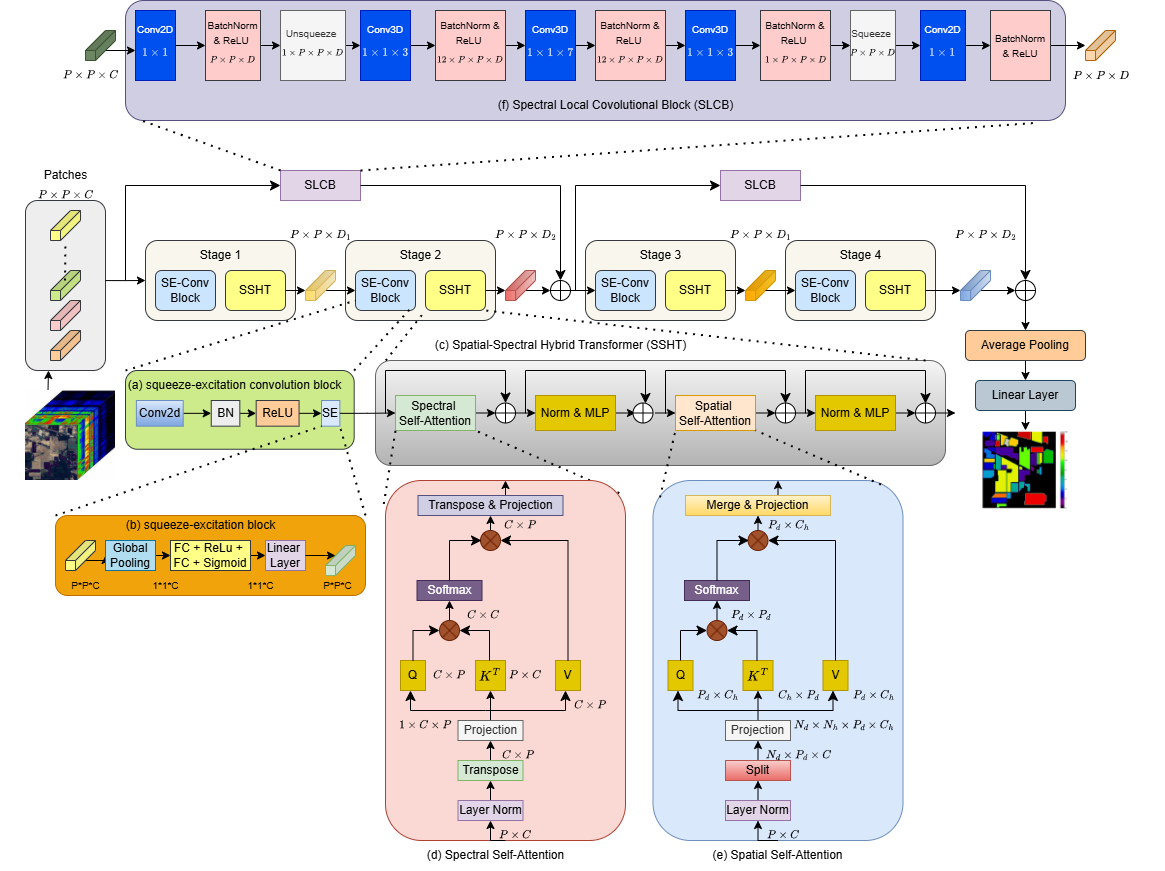}
			\caption{The framework of DATN \cite{SHU2024107351} is illustrated as an example of feature extraction in sequential/cascaded architectures. The HSI cube $X \in \mathbb{R}^{H \times W \times C}$, where $H$, $W$, and $C$ represent the spatial height, width, and spectral bands, respectively, is partitioned into 3D patches of size $P \times P \times C$. These HSI image patches will be processed by four stages of hierarchical blocks, where each stage contains an (a) squeeze-excitation-convolutional (SE-Conv) block and a (c) spatial-spectral hybrid transformer (SSHT) block. The SE-conv block includes a 1$\times$1 2D convolution layer, a BatchNorm layer, a ReLU activation layer, and a (b) squeeze-excitation (SE) block to control the number of feature map channels and to carry out initial feature extraction for the subsequent SSHT block within each stage. Each (c) SSHT contains a (d) Spectral Self-Attention and a sequential (e) Spatial Self-Attention to extract spatial-spectral features separately. The output shapes of feature maps after each stage are $P \times P \times D_1$, $P \times P \times D_2$, $P \times P \times D_1$, and $P \times P \times D_2$, respectively. The (f) spectral local-convolutional block (SLCB), which consists of sequential convolutional layers, unsqueeze, and squeeze operations, is introduced to emphasize the local spectral information. The kernel sizes of the convolutional layers and output sizes are denoted on the (f) SLCB module. The enhanced spectral features extracted by (f) SLCB are integrated with the output feature maps of SSHT through skip-connection.  }
			\label{datn}
		\end{figure}
		
		Most works used convolutional and pooling operations along different dimensions to extract spatial and spectral features separately and process these features with spatial attention and spectral attention. For instance, MATNet \cite{Zhang2023MATNeT} processed the input features sequentially by channel attention (CA), which consisted of the average-pooling layer and the max-pooling layer, a 2D convolution layer. And spatial attention (SA) was also composed by average-pooling and max-pooling operations with a convolutional layer. Then, these tokens were sent to cascaded Transformer encoders. S3FFT \cite{Xie2023fusion} first used 2D convolution-based spatial attention to capture spatial relationships and designed spectral attention, which contained global average-pooling and 1D convolution, to enhance spectral features. SSFT \cite{Qiao2023sstf} adopted a 3D convolutional layer to capture features across the spectral dimension and the 2D convolutional layer to capture spatial features. MSSTT \cite{electronics12183879} first utilized a 3D convolution-based Inception module for spatial-spectral information enhancement and a 2D convolution-based transposed Inception for spatial information enhancement. 
		
		To extract multi-granularity features, MCAL \cite{Xu2023multiscale} developed the Spectral Attention Module (SAM), which consisted of a global average-pooling layer and a global max-pooling layer to recalibrate spectral bands adaptively. Then, it hierarchically conducted multiscale feature extraction to model low-level, mid-level, and high-level features by combining the CNN and Transformer. GTFN \cite{Yang2023gtfn} combined the Graph convolutional network (GCN) for spatial modeling and the Transformer for spectral sequence modeling. CentralFormer \cite{Li2024CentralFormer} utilized parallel 3D convolutional layers with kernel sizes of 1$\times$1$\times$7 and 3$\times$3$\times$1 to extract spatial-spectral features and integrated them using another 3D convolution with a 1$\times$1$\times$1 kernel.
		
		In order to reduce the computational complexity of self-attention, LiT \cite{Zhang2023lightweight} deployed successive max-pooling layers to decrease feature dimension for position attention to extract local features and subsequent channel attention to model long-range dependencies. PASSNet \cite{Ji2023passnet} contained two convolutional blocks to extract local features and two Transformer blocks to extract and blend local–global spatial and spectral features. The convolutional blocks implemented spatial convolution, which only performed convolution on the first part of the input feature cube that was divided along the channel dimension.

		\item As shown in Fig \ref{pct}, numerous works for HSI classification utilized the \textbf{parallel CNN and Transformer branch} dual-branch architecture, which consists of a CNN branch to extract spatial/local features and a Transformer branch to learn global relationships \cite{rs14205251, app13010492, Li2023cnntrinter, rs15133269, Peng2023crossdomain, Yang2023multilevelIT, Cheng2024CACFTNet}. To be more specific, STransFuse \cite{Gao2021stransfuse} combined the Swin Transformer for global semantic information learning and the pre-trained ResNet34 network to extract the spatial contextual information in parallel. Inside the Inception Transformer block \cite{rs14194866}, the input features were first segmented proportionally along the channel dimension. Then, they were fed into two parallel branches, where one mainly contained max-pooling and parallel convolution operations, and the other mainly performed self-attention. 
		
		Furthermore, S3FFT \cite{Xie2023fusion} used MHSA to learn long-range global correlations and CNN for local correlations in the C-Transformer module. FusionNet \cite{rs14164066} also devised convolution modules as the local branch and convolution-Transformer modules for the global branch. MSSTT \cite{Meng2024msstt} adopted various convolutional kernels to extract multiscale local features and a multiscale super token attention (MSSTA) branch to capture global features. DCTN \cite{Zhou2024dctn}, which also consisted of a CNN branch and a Transformer branch, utilized three parallel 1$\times$1 2D convolutional blocks to extract features from height, width, and spectral dimensions within the Transformer branch. DBSSAN \cite{Zhao2024dbssan} developed a Transformer-based spectral branch to process spectral pixels and a spatial feature extractor, which implemented cosine similarity and Gaussian-Euclidean similarity to investigate the relationship between the central pixel and the neighboring pixels.
		
		As for the operations to extract features, 2D and 3D convolutions were commonly adopted by these models. After extracting global spectral information with 3$\times$3 group convolutions and spatial information with standard 3$\times$3 convolutions in parallel, CTMixer \cite{Zhang2022mixer} further used a Transformer encoder with a convolution (TEC) branch to learn global information and the CNN branch to model local features, as shown in Fig. \ref{ctmixer}. Inside the dual-channel block in \cite{rs14143426}, spatial features were extracted by three successive hybrid convolution and Transformer encoder blocks, and spectral features were modeled by 2D convolution and group-wise self-attention blocks. CTFSN \cite{Zhao2023fusionsplicing} devised a local information extraction branch, which is mainly composed of 3$\times$3 depth-wise convolution and a parallel global context information extraction branch that also used 3$\times$3 depth-wise convolution to process the output of self-attention. DCTransformer \cite{app14051701} adopted the channel attention mechanism (CAM) with discrete cosine transform (DCT) convolutional kernels to extract high-frequency spectral features. LGGNet \cite{Zhang2023LGGNet} fed the obtained features into the 3D spatial-spectral residual module, which is mainly composed of successive 3$\times$3$\times$1 3D spatial blocks and 1$\times$1$\times$3 3D spectral blocks, and the Transformer module in parallel. 
		
		Moreover, some works aimed to learn multiscale spatial or spectral features in this dual-branch architecture. FUST \cite{Zeng2023microscopic} used the CNN branch to extract multiscale spatial features and the parallel Transformer branch to highlight the critical spectral information. HCVN \cite{Yan2023hybridconvvit} used a spatial feature extraction branch, consisting of several cascaded HCV (hybrid convolution and ViT) modules, to obtain the deep global and local spatial features, and a parallel spectral feature extraction branch, including cascaded 3D CNN of different kernel sizes, to extract multiscale spectral features. SLA-NET \cite{Zhang2023morpho} combined spatial extractor and morphological extractor in parallel, where the spatial extractor also included two parallel convolution blocks and the morphological extractor contained multiple convolutions and Transformer blocks. 
		
		\begin{figure}
			\centering
			\includegraphics[scale=0.5]{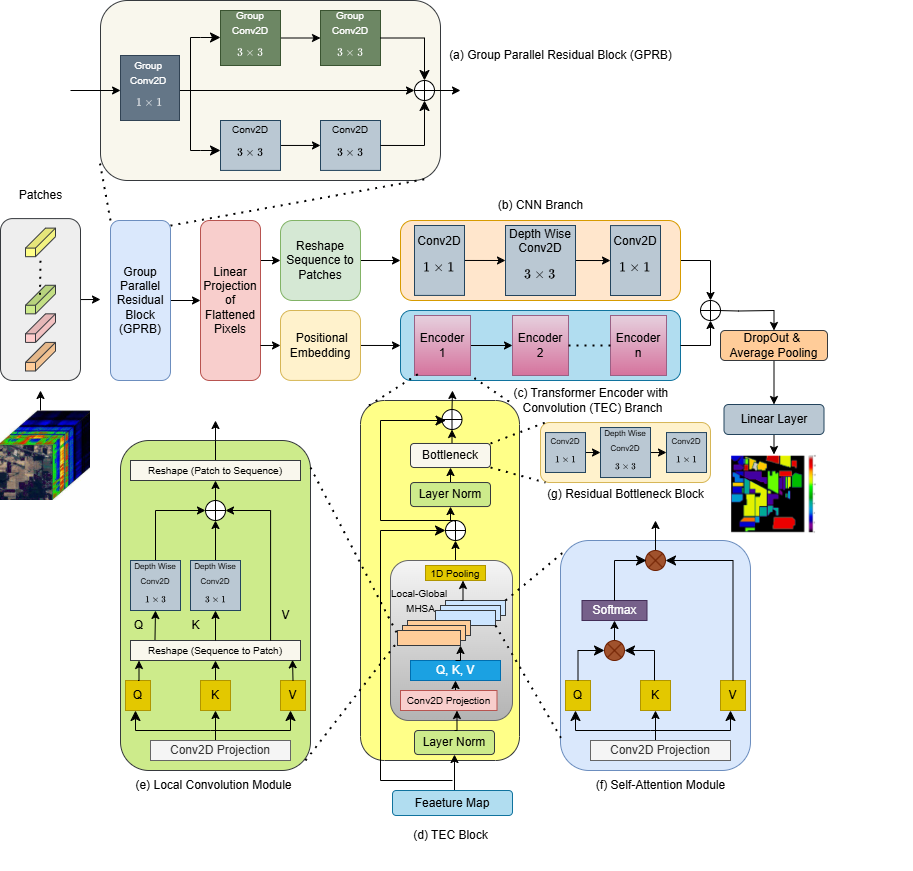}
			\caption{The overview framework of CTMixer \cite{Zhang2022mixer} is depicted as an example of parallel CNN and Transformer branches. The HSI hypercube $X \in \mathbb{R}^{H \times W \times C}$, where $H$, $W$, and $C$ represent the spatial height, width, and spectral bands, respectively, is divided into $N$ patches. The HSI patches are first processed by (a) Group Parallel Residual Block (GPRB), which contains parallel 2D convolutional layers, to extract grouped spectral information. The output feature map is linearly projected and flattened. The flattened feature map is reshaped to patches as inputs to the (b) CNN branch to extract spatial features. The positional embedding is added to the linearly mapped sequence vectors as inputs to the (c) Transformer Encoder with Convolution (TEC) branch, which contains $n$ TEC encoder blocks. The (d) TEC block adopts the Local-Global MHSA mechanism, which projects the feature map with 2D convolution to generate $Q$, $K$, and $V$ vectors. Half of the vectors are sent into (e) the Local Convolutional Module to extract local features, and the other half is fed to (f) the Self-Attention Module to capture global features effectively.}
			\label{ctmixer}
		\end{figure}
		
		\item Multiple works also adopt \textbf{Parallel Transformer branches}, consisting of a spatial Transformer and spectral Transformer to extract spatial-spectral features in parallel, as shown in Fig. \ref{ptt}. The spatial and spectral features are discriminated by dividing or convolution operations along spatial and spectral dimensions separately, such as Hyper-ES2T \cite{WANG2022103005}, BS2T \cite{Song2022bs2t}, CESSUT \cite{Xin2022conhsi}, S2Former \cite{electronics12183937}, CS2DT \cite{Xu2023cs2dt}, FTSCN \cite{Liang2023SimAMCNN}, GTCT \cite{Qi2023globallocal}, FactoFormer \cite{Mohamed2023factorformer}, Dual-MTr \cite{Li2024DualMTr}, D$^2$BERT \cite{rs16030539}, and DBMST \cite{Shi2024dbmst}. For instance, DAFFN \cite{Cui2023DAFFN} and DBFFT \cite{Dang2023DoublebranchFF} achieved channel and spatial feature separation with depth-wise separable convolutions. TransCNN \cite{Hao2022featurefusion} used three CNN branches composed of 3D and 2D convolutions to extract features along the three dimensions of HSI cubes. Then, the features were processed by three parallel Transformer encoders. MS$^3$DT \cite{Pan2023dualtrans} segmented patches by channel and used a linear projection to generate channel tokens for spectral Transformer, while the parallel spatial Transformer module divided the features by row and column to generate row and column tokens that were processed using a double-branch MHSA separately. 
		
		MATA \cite{Liu2023multiarea} passed multi-area inputs corresponding to a target pixel into a multiscale target attention module (MSTAM) simultaneously, where the computations of multi-area inputs of a target pixel were parallel. HMSSF \cite{He2024hybridmultiscale} mapped the cropped pixel-wise patch cubes into the spatial token and spectral token, then processed them in the Multiscale Spatial Transformer Branch and Multiscale Spectral Transformer Branch, which independently extracted multiscale spatial and multiscale spectral representations, respectively. 
		
		Furthermore, as previously discussed, some works used HSI patches for spatial information extraction and pixels for spectral information learning \cite{He2022twobranch, Lu2023multiattention, Kumar2023methanemapper}. For instance, DSS-TRM \cite{Liu2022dsstrm} converted image blocks in each band into 1D feature vectors, arranged in band order, and composed image blocks for spatial Transformer using three principal components. In S$^2$FTNet \cite{Liao2023fusion}, the spectral Transformer module linearly mapped the HSI pixel for MHSA, while the spatial Transformer module contained successive 3D convolution, 2D convolution, and MHSA. Then, three spatial Transformer blocks, which used different pooling operations, were used in parallel to explore the multiscale long-distance dependency of images.  
		
		\item Additionally, some works separately extracted various features with \textbf{multiple parallel modules and subsequent cascaded Transformer} blocks to model the global relationships \cite{Tang2023double, Zhang2023SAnet, Li2024CreatingNet}, as depicted in Fig \ref{pcct}. Usually, the parallel feature extraction modules mainly consist of different convolutional layers. More specifically, SpectralSWIN \cite{Selen2022SpectralSWIN} and NEHT \cite{rs14194732} extracted spatial features with 2D convolutional layers and spatial-spectral features with parallel 3D convolution blocks. D$^2$S$^2$BoT \cite{Zhang2023D2S2BoT} used 2D convolution blocks to extract spatial features and parallel 3D convolution blocks to extract spectral features. The extracted features were then passed to Dual-Dimension Spectral-Spatial Bottleneck Transformer encoders, including channel global attention and spatial global attention. 
		
		Furthermore, MST-SSSNet \cite{Gao2024MST-SSSNet} utilized parallel 2D convolution layers to extract spatial-spectral feature maps, which were flattened and tokenized for the subsequent main–sub Transformer module. H2MWSTNet \cite{Zhong2024h2mwstnet} was divided into three phases; each phase contained a dual-branch spatial-spectral convolution (DBSSC) and several multi-granularity window shift Transformer (MWSFormer) encoder blocks. The DBSSC consisted of a parallel spatial convolutional module and a spectral convolution module. The parallel local and global branches of HiT \cite{Yang2022hsiT}, which are mainly composed of 3D convolutions, were developed to capture local spatial-spectral features and long-range spectral information. The aggregated features were further processed by cascaded Conv-Permutator modules, which mainly contained three parallel 1$\times$1 2D convolutions along the three dimensions of HSI cubes. 
		
		In addition, morphFormer \cite{Roy2023morpho} deployed parallel spatial and spectral morphological blocks, each consisting of parallel branches of dilation and erosion, followed by 2D convolutions of different kernel sizes (3$\times$3 for spatial and 1$\times$1 for spectral kernel), as shown in Fig. \ref{morphformer}. A dilated image is obtained by selecting the pixel with the maximum value, while erotic images are acquired by selecting the pixel with the minimum value. CASST \cite{Peng2022crossattention} used different convolutions for fine-grained spatial features and grouped spectral sequence information. Two parallel 3D convolutions with different kernels (1$\times$1$\times$7 and 3$\times$3$\times$7) were used to extract spectral and spatial feature maps \cite{rs14153705}. Besides, two orthogonal 2D convolutions with kernel sizes 3$\times$1$\times$3 and 3$\times$3$\times$1 in parallel were used to replace the 3D convolution of kernel size 3$\times$3$\times$3 to learn spatial-spectral features in MST-SSSNet \cite{Gao2023MSTSSSNet}. 
		
		In order to learn multiscale features, MHCFormer \cite{Shi2023mhcformer} designed a multiscale spatial branch and multiscale spectral branch in parallel, where the spatial branch contained three 3D convolution blocks of different kernel sizes in parallel (3$\times$3$\times$1, 5$\times$5$\times$1, and 7$\times$7$\times$1). The spectral branch also included three 3D convolution blocks in parallel (1$\times$1$\times$3, 1$\times$1$\times$5, 1$\times$1$\times$7). SS-TMNet \cite{rs15051206} also deployed parallel 3D convolutions with different kernel sizes to learn multiscale spatial-spectral features first and then used three 2D convolutions to perform height, width, and spectral attention in parallel. 
		
		Moreover, ELS2T \cite{Zhang2023ELS2T} first utilized parallel 3D CNN blocks with different kernel sizes to learn spatial and spectral features independently, then used three different dilate convolutions to generate multiscale features. The information was aggregated along the two spatial directions with horizontal average-pooling, horizontal max-pooling, vertical average-pooling, and vertical max-pooling, respectively. PyFormer \cite{Ahmad2024PyFormer} developed parallel 3D convolutions to obtain multiscale feature maps by sequential downsampling and upsampling on the spatial dimensions. The output of these pyramid levels was concatenated as the input tensor for the Transformer block.
		
		Furthermore, MAR-LWFormer \cite{Fang2023multiattention} extracted three different types of features, i.e., multiscale Spectral–Spatial, extended morphological attribute profile (EMAP), and local binary pattern (LBP) features of HSI with three parallel branches, where spatial-spectral features were learned with 3D convolutions of different kernel sizes, EMAP features can capture the edge and texture information, while LBP achieves the extraction of spatial information of the whole image by describing the differences between each central pixel and neighboring pixels. Additionally, MT-CW \cite{Tulapurkar2022mha} designed a CNN feature extractor (CFE), which used 3D and 2D convolutions to extract spatial and spectral features and a parallel Wavelet decomposition module (WM) to decompose HSI images into various approximations using Coiflet wavelet transform.   
		
		\begin{figure}
			\centering
			\includegraphics[scale=0.5]{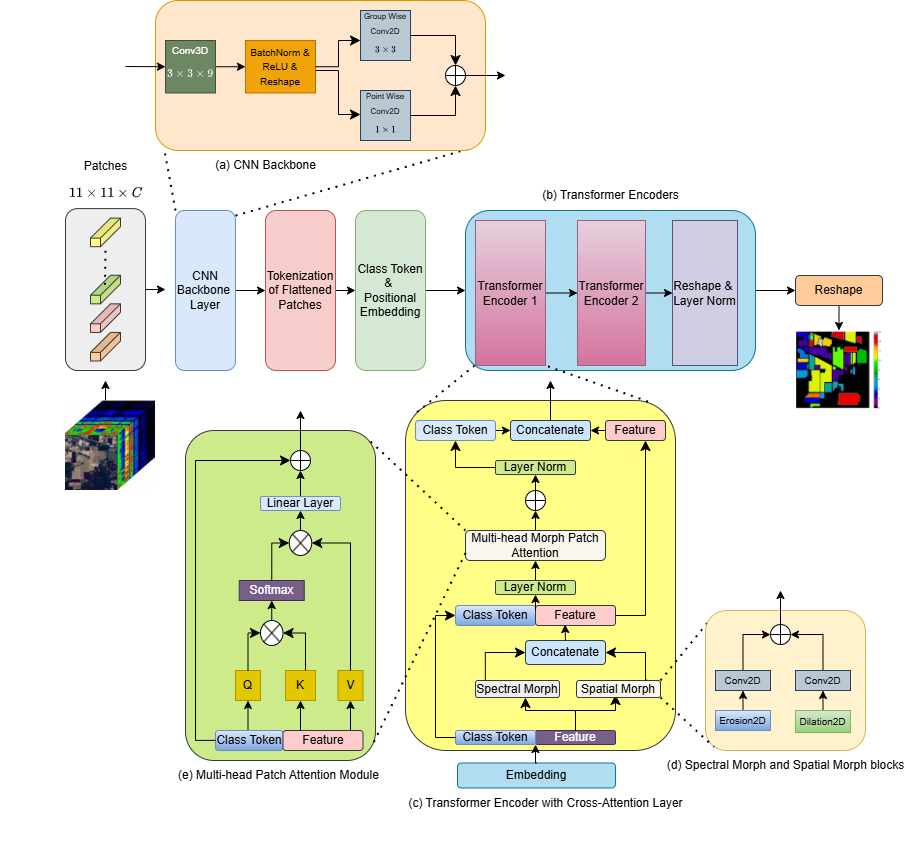}
			\caption{The overall framework of morphFormer \cite{Roy2023morpho} is illustrated as an example of multiple parallel modules and subsequent cascaded Transformer architectures. The HSI hypercube $X \in \mathbb{R}^{H \times W \times C}$, where $H$, $W$, and $C$ represent the spatial height, width, and spectral bands, respectively, is divided into $11 \time 11 \times C$ subcubes. The (a) CNN backbone, which consists of a 3D convolutional layer and parallel 2D convolutional layers (group-wise and point-wise), is applied to reduce spectral dimension and extract abstract features. Then, the extracted feature is tokenized with the tokenization module from SSFTT \cite{Sun2022tokenization}. The Gaussian weighted tokenized features with positional embedding and class tokens are passed into cascaded (b) Transformer encoders, which consist of multiple (c) Transformer Encoder with Cross-Attention Layer. The feature map of the tokenized embedding is fed to (d) a spatial morphological (SpatialMorph) block and a spectral morphological (SpectralMorph) block. The morphological blocks contain parallel erosion, dilation, and convolutional layers. The class tokens are used to exchange information between HSI patches within (e) Multi-head Patch Attention Module.}
			\label{morphformer}
		\end{figure} 
		
		\item \textbf{Other architectures}: 
		Some works integrated Transformer with different RNN networks in varying schemes. GBiLSTM-MFCT \cite{Xu2022biLSTM} consists of a GBiLSTM branch that extracted the group-wise spectral features with bi-directional LSTM and an MFCT branch with a multi-stage fusion convolutional Transformer. HybridGT \cite{Neela01022025} integrated LSTM and Graph Transformer in parallel for feature selection and utilized SVM as the classifier. SRT \cite{Zhou2024srt} integrated RNN and Transformer by computing self-attention within the RNN structure. The current hidden state was updated by summing the previous hidden state with the self-attention features, which were generated by calculating $Q$, $K$, and $V$ with the previous hidden state and input. S3L \cite{rs16060970} incorporated a GRU with a Transformer to emphasize the sequential dependence of spectral features. 
		
		Moreover, HSI-TNT \cite{Liu2022coastal} employed a \textbf{Transformer iN Transformer} architecture, where the inner Transformer block was used to establish the relationship between pixels to extract local features, and the outer Transformer block was used to establish the relationship between patches to extract global features. In addition, HSI-TransUNet \cite{NIU2022107297} formed the feature extraction and spectral feature attention blocks with a UNet design, U$^2$ConvFormer adopted a nested U-Net to extract and aggregate multiscale spatial-spectral features, and DiffSpectralNet \cite{Sigger2024DiffSpectralNet} implemented the U-Net denoising network to extract spatial-spectral features, which were learned in the unsupervised diffusion process. RS-Net \cite{electronics13204046} employed a Random Forest classifier to aggregate the prediction results from the SpectralFormer network.
		
	\end{enumerate}
	
	According to experimental results, the above architectures can improve classification performance with branches that aim to enhance different feature extraction. However, the computational complexity increased dramatically with these multi-branch or cascaded architectures. For instance, MATA \cite{Liu2023multiarea} proposed to extract features from each scale with a different MHSA module. Moreover, 2D and 3D convolution layers were broadly adopted with Transformer modules to complement the extraction of local features, resulting in more model parameters and training time. Besides, the separately extracted features require additional modules for better aggregation, which usually demands more complex operations. The feature fusion will be discussed in Section \ref{featurefusion}.

	\subsection{HSI Feature fusion}\label{featurefusion}
	
	After separately extracting features from HSI data, feature fusion/aggregation is essential for further processing. One approach is \textbf{addition}: features captured by different branches achieve feature fusion through element-wise addition \cite{Zhang2022mixer, rs15051206, Qian2023neighborhood}, channel addition \cite{Yu2022mstnet}, addition through residual connection \cite{LI2024123939}, and spatial dimension addition \cite{rs14194732}. Some works only added class tokens to gain a global representation \cite{Zhou2022hussat, YANG2023145}. 
	
	Besides, \textbf{concatenation} is the most commonly applied method to fuse different tokens, including positional tokens and class tokens with features \cite{Zhou2022rethinking}, inverse sequences from bi-directional RNN \cite{Zhao2023rnntrans}, and features from different branches \cite{Xin2022conhsi, rs14143426, He2022CSiT, He2022twobranch, Liang2023SimAMCNN, Lu2023multiattention, Xu2023cs2dt, Pan2023dualtrans}, levels \cite{Yu2022mstnet}, or scales \cite{Wu2023tokenhash}. Moreover, the input features to Transformers can also be selected according to the correlation between the target pixels and neighboring pixels based on the adjacency matrix, and the top points after sorting can be kept and concatenated (feature maps' spatial information is aggregated through average-pooling and max-pooling) \cite{Yang2023gtfn}. Then, extra operations can further process the concatenated features, including linear projection, convolution, layer normalization, and activation functions for classification \cite{Xu2022biLSTM, Pan2023dualtrans, Zhao2023fusionsplicing}. 
	
	Concatenation and addition can also be applied to different components within the same network \cite{Li2023cnntrinter}. However, some researchers found that concatenating positional embeddings with token embeddings achieves better results than adding. It is possible that the concatenating ensures the integrity of multiscale positional information and avoids the confusion between positional and semantic information \cite{Wu2023tokenhash}. 
	
	Furthermore, \textbf{convolution} operation is also widely applied for feature fusion. Some works used convolutions to reduce the dimension of the fused features for lower computational complexity \cite{rs14205251, Yang2022hsiT, Chen2023projection, Zhang2023ELS2T}, while others used convolutional layers to fuse multiscale or multi-level features \cite{Cui2023DAFFN, Zhang2023SAnet, Shi2023mhcformer, Yang2023multilevelIT, Zhang2023MATNeT}. Moreover, STransFuse \cite{Gao2021stransfuse} added the average-pooling layer after concatenation and convolution operations to process the fused features for further decreasing computational complexity. FUST \cite{Zeng2023microscopic} also adopted dimensional concatenation, 3$\times$3 convolution, and global average-pooling to fuse features from parallel CNN and Transformer branches. 
	
	The feature fusion can also adopt a \textbf{weighted} fashion, where the outputs of different branches are fused with a tradeoff parameter $\alpha$ to balance the importance of different features as $f =\alpha \mathbf {F}_{\textbf{spatial}} + \left ({1-\alpha }\right) \mathbf {F}_{\textbf{spectral}}$, or with learnable weights, such as $f = \mathbf {W}_{\text {spa}} \odot \mathbf {F}_{\text {spatial}} + \mathbf {W}_{\text {spe}} \odot \mathbf {F}_{\text {spectral}}$ and $f = \textbf{concat}[\mathbf {W}_{\text {spa}} \odot \mathbf {F}_{\text {spatial}}; \mathbf {W}_{\text {spe}} \odot \mathbf {F}_{\text {spectral}}]$, \cite{Zhao2022crossscene, Zhang2023D2S2BoT, Zhang2023morpho, app13010492, Liao2023fusion, Qi2023globallocal}. Besides these works, soft attention weights were employed to balance the spectral and spatial feature \cite{rs14153705}. MFSwin-Transformer \cite{agronomy12081843} assigned different weights to the features from different stages. The feature weight was multiplied with the feature vector element by element, and finally, the multiscale fusion feature was obtained. Considering that different object types have different feature preferences, SCSTIN \cite{rs15133269} imposed weights on each object type's spatial and spectral features. RSAGformer \cite{Zu2023wrsag} used a weighted residual connection to fuse the attention scores between two adjacent layers. Many works also fused the different features through the attention mechanism, which will be discussed in Section \ref{mhsa}. 
	
	Some researchers devoted efforts to \textbf{comparing} the different feature fusion methods. For instance, AttentionHSI \cite{rs14091968} compared three aggregation techniques: add the spatial and spectral attention scores, concatenate the spatial and the spectral attention feature map along channel dimension, and Hadamard product between the spatial and spectral attention matrix with values ($V$). DSS-TRM \cite{Liu2022dsstrm} compared concatenation, point-wise addition, and point-wise multiplication of spatial and spectral features. Moreover, MATA \cite{Liu2023multiarea} compared the multiscale information interaction methods, including equally treated approaches (concatenation, adding, and pooling) and weighting approaches with trainable weights for multiscale feature fusion. 
	
	In order to enhance the information interaction, the information fusing from different branches (usually spatial and spectral) can be achieved through exchanging and concatenating class tokens from other branches \cite{Peng2022crossattention, Roy2023morpho, Dang2023DoublebranchFF, He2022CSiT, Xu2023cs2dt}, as shown in Fig. \ref{morphformer}, which depicted the structure of morphFormer \cite{Roy2023morpho} as an example. TECCNet \cite{PAN2024106973} added class tokens from spectral and spatial tokens, and the newly generated tokens were $Q$, $K$, and $V$ for self-attention. APSFFT \cite{Huang2024APSFFT} can also utilize the class token from another branch to calculate the $Q$ matrix for self-attention. 
	
	Additionally, features can also be enhanced by splicing a weighted sum of similarity values among each feature element \cite{Zhou2023dictionary}. Channel Shuffle \cite{Li2023spformer} allows channel information to flow between different groups without increasing any parameters. FusionNet \cite{rs14164066} designed a down module and an up module, which used 1$\times$1 convolution for dimension alignment to exchange information between local and global branches. In addition, LGGNet \cite{Zhang2023LGGNet} performed information interaction between two branches through reshaping and addition. CAF-Former \cite{Xu2024CAF-Former} proposed a cross-attention fusion module to fuse the features extracted by the convolution branch and the Transformer branch by exchanging the Key ($K$) matrix. HyperSINet \cite{Yu2024hypersinet} established a synergetic interaction network to exchange information between the Transformer branch and CNN branch using weighted linear summation.
	
	These feature fusion modules were mainly the consequence of separate feature extraction. The extra feature fusion module further increases these Transformer-based networks' computational cost and memory burdens. Moreover, the proper design of feature aggregation modules is also essential for classification performance. Research works that utilized loss functions to fuse features from different branches will be introduced in Section \ref{loss}.

	\section{Multi-head Self-attention}\label{mhsa}
\begin{figure}
		\centering
		\includegraphics[scale=0.4]{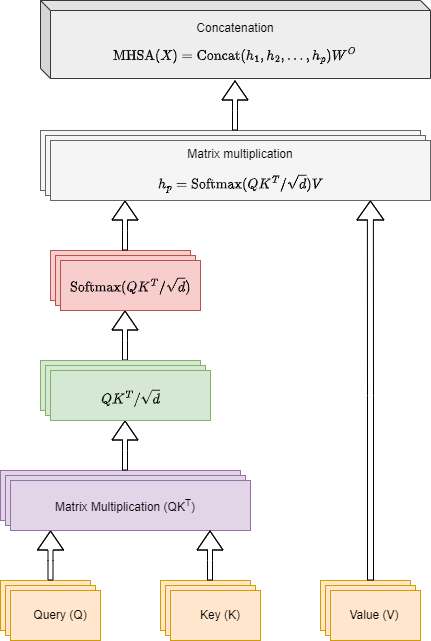}
		\caption{The scaled multi-head self-attention \cite{Vaswani2017attention}. }
		\label{fig:mhsa}
	\end{figure}

	The original scaled dot-product attention is formulated as \cite{Vaswani2017attention}:
	
	\begin{equation} 
		\text {Attention}(Q, K, V)=\text {Softmax}\left ({\frac {QK^\top }{\sqrt {d_{k}}}}\right)V
	\end{equation}
	
	\noindent where $Q$, $K$, and $V$ represent the query, key, and value, respectively, and $d$ is the input data dimension. The MHSA uses $p$ heads $(h_{1}, h_{2},\ldots,h_{p})$, which can be written as: 
	
	\begin{equation} 
		h_{p} = \text {Attention}\big (XW_{p}^{Q}, XW_{p}^{K}, XW_{p}^{V}\big)
	\end{equation}
	
	\noindent with projections using learned parameter matrices $W_{p}^{Q}, W_{p}^{K}, W_{p}^{V}$. Different heads of the MHSA mechanism learn different attentions independently and in parallel. The MHSA is obtained by concatenating heads to form a larger feature matrix, as:
	
	\begin{equation} 
		\text {MHSA}(X) = \text {Concat}(h_{1}, h_{2}, \ldots,h_{p})W^{O}
	\end{equation}
	
	\noindent where $W^O$ represents the learned parameter matrices on the concatenated heads. The calculation process is shown in Fig \ref{fig:mhsa}. 
	
	The spectral bands of HSI data contain abundant information. Thus, paying more attention to the crucial spectral channels can enhance the performance of the HSI classification tasks. Moreover, because the adjacent spectral bands are highly correlated with each other and non-adjacent spectral bands also demonstrate long-term dependency \cite{ZHOU201939lstm}, some researchers designed Transformer modules to focus on the spectral information by applying self-attention on the projected features of spectral bands \cite{Zhou2021swinspectral, Zhou2022hussat}. 
	
	For instance, MSBMSW \cite{Li2023shift} developed a two-stage self-attention to enhance the spectral information. In the first stage, the self-attention mechanism was used to acquire sequence relations of the $l$ spectral band. In the second stage, a window shift with a stride of $l/2$ was designed to exchange information between bands by performing self-attention for the second time. In addition, Grid-Transformer \cite{Guo2023gridfewshot} splits $Q$, $K$, and $V$ into multiple heads along the spectral channel dimension. After computing the self-attention, head features are concatenated into a spectral-wise enhanced feature map. Since neighboring pixels were generally more crucial than those far from the target pixel, DiCT \cite{Zhou2023dictionary} generated $Q$ and $K$ by grouping neighboring pixels to reduce interference from anomalous elements. 
	
	Furthermore, as discussed in Section \ref{featureextract}, various forms of Transformer blocks were widely applied to extract spatial-spectral features \cite{Xie2023fusion}, and numerous variations of self-attention mechanisms were introduced for different purposes. Therefore, the following part of this section will briefly review the modifications of the self-attention mechanism from various perspectives.

	\subsection{Generating $Q$, $K$, and $V$}
	
	Conventionally, the $Q$ (query), $K$ (key), and $V$ (value) were generated by linear projecting the input features, while other operations can also be applied to generate the $Q$, $K$, and $V$ for self-attention computation, including convolutions \cite{Zhang2023D2S2BoT, Qi2023globallocal, Lu2023multiattention, Qian2023neighborhood, Zhang2022mixer}, pooling \cite{Kang2023ssloil, Li2023spformer, Zhang2023lightweight}, or combinations of linear, convolution, and pooling \cite{Song2022bs2t, Bai2022multibranch, Xu2022biLSTM, Zhao2023fusionsplicing}. For instance, S2Former \cite{electronics12183937} projected spatial context features into $Q$, $K$, $V$ by applying 1$\times$1 point-wise convolutions and 3$\times$3 depth-wise convolutions in a spectral-wise manner. HybridFormer \cite{Ouyang2023hybrid} contained spatial attention and spectral attention, both of which were computed as $Conv2D(QK^T)V$. Moreover, Hyper-ES2T \cite{WANG2022103005} used a linear projection to generate $Q$, while the combination of a 1D convolution layer and a linear projection layer was utilized to obtain $K$ and $V$. PASSNet \cite{Ji2023passnet} generated $Q$ with linear projection, and $K$ and $V$ with Patch Attention Module, which mainly consisted of average-pooling kernels along channel in the horizontal and vertical directions and point-wise convolutions. STransFuse \cite{Gao2021stransfuse} calculated $Q$ with convolutions while output $K$ and $V$ using adaptive 2D average-pooling. CITNet \cite{Liao2023integrated} linearly mapped $Q$, $K$, and $V$, and different attention heads were concatenated and fused with the $V$ after convolution to obtain the CMHSA output, as $\text {CMHSA}(X) = \text {Concat}(h_{1}, h_{2}, \ldots,h_{p})W^{O} + Conv(V)$. The Outlook Attention in MAT-ASSAL \cite{Zhao2023multiattention} directly generated a matrix via linear layers of weights rather than the matrix multiplication between $Q$ and $K$. MSSTT \cite{electronics12183879} used sine functions to regularize the attention output values that fall within the effective range of the activation function due to the periodicity of sine. 
	
	There are also works that adopted multiscale matrices for self-attention mechanism. For example, MSST \cite{rs16020404} generated keys and values of different scales based on downsampled multiscale tokens, and utilized convolution and linear projection to fuse multiscal $K$ and $V$, while TNCCA \cite{rs16071180} obtained multiscale $Q$, $K$, and $V$ from the same token with different 2D convolutional kernels (kernel sizes of 3$\times$3 and padding of 1 for $Q$, kernel sizes of 5$\times$5 and padding of 2 for $K$, and kernel sizes of 3$\times$3 and padding of 2 for $V$).
	
	These operations to generate $Q$, $K$, and $V$ for self-attention can be considered additional shallow feature extraction. Although convolutions can further extract the structural information from the input embeddings, the computational burden grows compared to linear projection and pooling layers. However, the simple linear mapping and pooling may lose some critical information about the input embeddings. Balancing the tradeoff between fully extracting features and computational complexity is an essential topic while designing networks.
	
	\subsection{Feature fusion in self-attention}
	
	\begin{figure}
		\begin{subfigure}{0.23\textwidth}
			\hspace{-0.3in}
			\includegraphics[scale=0.25]{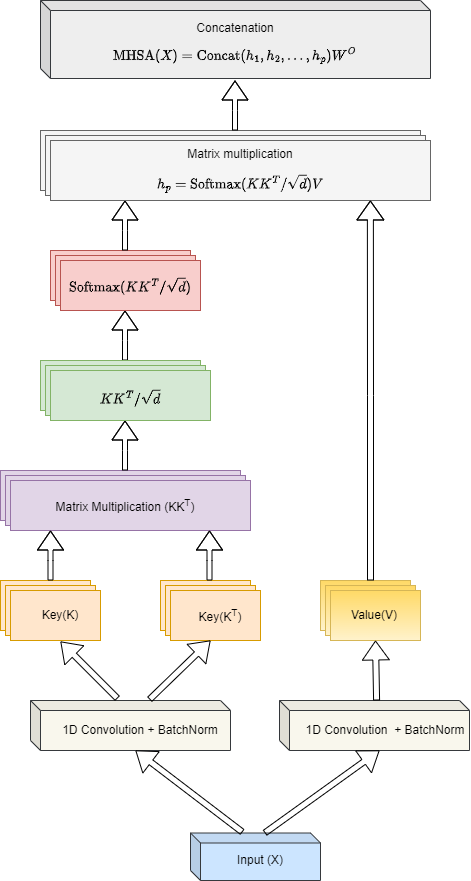}
			\caption{}
			\label{mata}
		\end{subfigure}
		\hfill
		\begin{subfigure}{0.23\textwidth}
			\hspace{-0.3in}
			\includegraphics[scale=0.25]{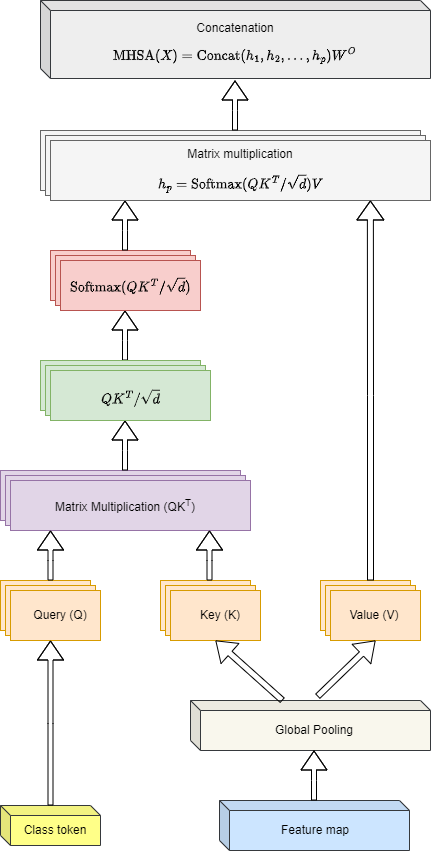}
			\caption{}
			\label{lgg}
		\end{subfigure}
		\hfill
		\begin{subfigure}{0.23\textwidth}
			\hspace{-0.3in}
			\includegraphics[scale=0.25]{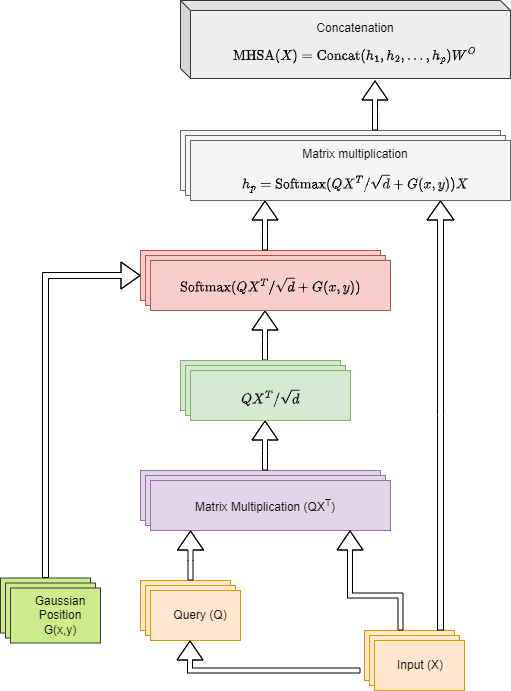}
			\caption{}
			\label{lgsa}
		\end{subfigure}
		\hfill
		\begin{subfigure}{0.23\textwidth}
			\hspace{0.4in}
			\includegraphics[scale=0.25]{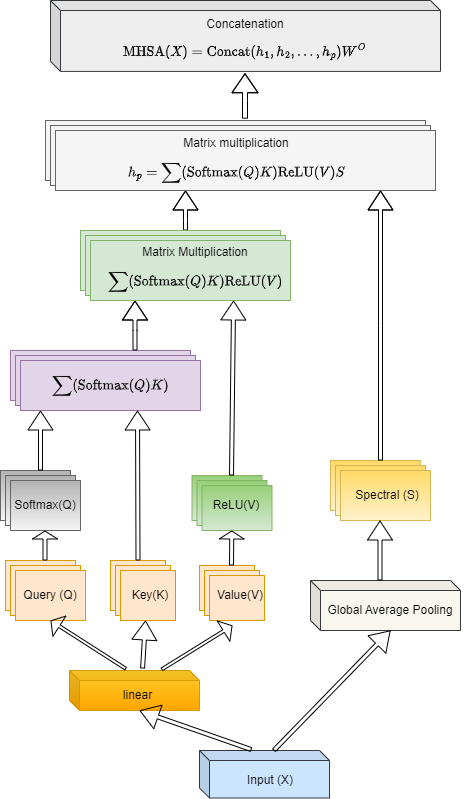}
			\caption{}
			\label{els}
		\end{subfigure}
	\caption{An illustration of the self-attention module of network: (a) MATA \cite{Liu2023multiarea}; (b) LGGNet \cite{Zhang2023LGGNet}; (c) LSGA-VIT \cite{Ma2023gaussian}; (d) ELS2T \cite{Zhang2023ELS2T} }
	\end{figure}
	
	The classic self-attention utilized the same input token to generate $Q$, $K$, and $V$ by multiplying a learnable weight. However, many works achieved feature fusing by applying different token embeddings to generate $Q$, $K$, and $V$ in the same self-attention module \cite{Fang2023multiattention}. For instance, TransCNN \cite{Hao2022featurefusion} initialized $Q$ with feature maps from different parallel branches, which extracted hypercube features along the three dimensions. MS2I2Former \cite{Cheng2024MS2I2Former} exchanged the $Q$ from two feature maps, and TNCCA \cite{rs16071180} switched $V$ from two branches. CreatingNet \cite{Li2024CreatingNet} used $Q$ from one branch while $K$ and $V$ from another branch to calculate self-attention. LGGNet \cite{Zhang2023LGGNet} and morphFormer \cite{Roy2023morpho} used class tokens to calculate $Q$ and feature embeddings to obtain $K$ and $V$. An illustration of the self-attention module in LGGNet is shown in Fig. \ref{lgg} and the overall framework of morphFormer is depicted in Fig. \ref{morphformer}. 
	
	Moreover, CSIL \cite{YANG2023145} used the embeddings from the neighbor region in the middle as the $Q$ and extracted a portion of surrounding embeddings at multiple granularities as $K$ and $V$. In a cross-domain scenario, $Q$ is from the source domain, while $K$ and $V$ are from the target domains to align features with self-attention \cite{Ling2023cdvit}. SANet \cite{Zhang2023SAnet} obtained feature tokens in depth, height, and width dimensions simultaneously. With $Q$ generated from the token on the current position, $K$ and $V$ are acquired in all positions. DATE \cite{Tang2023double} concatenated the spectral tokens with the corresponding spatial tokens at the same band. In a graph-Transformer network, the node features of two linked nodes were applied to compute $K$ and $V$, while the edge feature was used for $Q$. CTIN \cite{Li2023cnntrinter} obtained $V$ by the cosine similarity between the center vector and other vectors, while $Q$ and $K$ were from input tokens. Directly using self-attention as a feature fusion module can alleviate the requirement of adding an extra feature aggregation module. Thus, the computation burden can be relieved. 
	
	Additionally, CACFTNet \cite{Cheng2024CACFTNet} calculated a correlation matrix by computing the product of regional Q$^{r}$ and K$^{r}$, which were obtained by averaging $Q$ and $K$ within each divided HSI region, and keep the top k correlation connections to obtain the correlation information between image regions. 
	
	Within the spectral branch, DAHIT \cite{Shi2024DAHIT} calculated multiscale features for $V$ with convolutional kernels of different sizes and diagonal masked correlation matrix between different spectral bands, which were obtained by multiplying the $Q$ and the $K$. Moreover, in order to avoid irrelevant connections between the target token and the ones outside the neighborhood, LESSFormer \cite{Zou2022LessFormer} employed an element-wise mask matrix to multiply the attention matrix as:
	
	\begin{equation}
		\text {Attention}(Q, K, V)= \mathrm {Softmax}\left (M \odot \frac{QK^T}{\sqrt{d}} \right)V
	\end{equation}
	
	\noindent where the mask matrix $\mathbf {M}$ had ones on the diagonal and adjacent superpixels. 
	
	Similarly, RNN-Transformer (RT) \cite{Zhao2023rnntrans} utilized spectral and spatial soft masks to mitigate the negative impacts from different labels within the same patch. The spectral soft mask was generated based on the Gaussian function by computing a spectral pair distance matrix with the Euclidean distance, where longer spectral distances received a lower weight. The spatial distance between two pixels was calculated based on spatial coordinates. The spectral and spatial soft masks are integrated into the self-attention as:
	
	\begin{equation}
		\text {Attention}(Q, K, V)= \textbf {Softmax}\left (\frac{QK^T}{\sqrt{d}} \odot \boldsymbol {M}_{\mathrm{ spe}} \odot \boldsymbol {M}_{\mathrm{ spa}}\right)V 
	\end{equation}
	
	\noindent Additionally, G2T \cite{Shi2023g2t} exploited the adjacency relationship among superpixel nodes by incorporating the diagonal degree matrix of the graph as:
	
	\begin{equation}
		{\text {Attention}}(Q, K, V) = \left ({\text {Softmax}\left (\frac {QK^{T}}{\sqrt {d}} \right)+\left ({\mathbf {D}^{-(1 / 2)} \tilde {\mathcal {E}} \mathbf {D}^{-(1 / 2)}}\right)}\right) \mathrm {V}
	\end{equation}
	
	\noindent where $\mathbf {D}=(d_{i i})=\sum _{j} \tilde {\mathcal {E}}_{i j}$ is the diagonal degree matrix of graph. Moreover, re-attention was introduced by using a learnable transition matrix $\Theta$ to map the attention score into a new regenerated map before multiplying with $V$ \cite{rs14153705, Yang2024CAMFT}, as: 
	
	\begin{equation}
		\text {Attention}(Q, K, V) = Norm \left(\Theta^T \left(\textbf {Softmax} \left( \frac{ QK^T}{ \sqrt{d}}\right)\right)\right)V
	\end{equation}
	
	Furthermore, CCSF-Transformer \cite{Zhu2024centercate} proposed the spectral-salient-focused attention module (SFA) to incorporate global dependencies and enhance the spectral feature representation by by calculating a similarity matrix ${r_{ij}} \in {{\mathbb{R}}^{b \times b}}$ between $q \in {{\mathbb{R}}^{b \times c}}$ and $k \in {{\mathbb{R}}^{b \times c}}$:
	
	\begin{equation}
		{r_{ij}} = \frac{{q \otimes {k^T}}}{{{d_k}}} = \frac{{q \otimes {k^T}}}{{Diag\left( {k \otimes {k^T}} \right)}}: = \left( {\begin{array}{ccc} {\frac{{{r_{11}}}}{{d_k^1}}}& \cdots &{\frac{{{r_{1c}}}}{{d_k^1}}} \\ \vdots & \ddots & \vdots \\ {\frac{{{r_{b1}}}}{{d_k^b}}}& \cdots &{\frac{{{r_{bc}}}}{{d_k^b}}} \end{array}} \right)
	\end{equation} 
	
	\noindent where $\bigotimes$ is the matrix multiplication. The recalibrated global spectral weight matrix is obtained by $w_{spectral}^{sfa} = \text {Softmax} \left( {{r_{ij}}} \right)v$. The output map of SFA module is $x_{output}^{sfa} = x_{input}^{sfa} + x_{input}^{sfa} \odot w_{spectral}^{sfa}$. 
	
	In addition, MHIAIFormer \cite{Kong2024MHIAIFormer} designed a Multihead Interacted Additive Self-Attention (MHIASA) module, which generates a single global query vector $q\ = \mathop \sum \limits_{i = 1}^N {{\alpha }_i}*{{Q}_i}$, where the global attention query vector $\alpha \in {{R}^D}$ is calculated as: 
	
	\begin{equation} 
		{\alpha \ = \frac{{\exp \left( {{{Q}^T}\ \cdot \frac{{{{w}_a}}}{{\sqrt{N} }}} \right)}}{{\mathop \sum \nolimits_{j = 1}^D \exp \left( {{{Q}^T}\ \cdot \frac{{{{w}_a}}}{{\sqrt{N} }}} \right)}}} 
	\end{equation}
	
	\noindent where ${{w}_a} \in {{R}^N}$ is a learnable parameter vector, $N$ is the length of the token sequence and $D$ is embedding dimension.

	\subsection{Lower computational cost}
	
	The computational cost of self-attention mechanisms is quadratic to the input feature. Thus, researchers proposed various approaches to reduce the computational cost. Some of them aimed to decrease the dimensions of input features. For instance, LSFAT \cite{Tu2022aggregation} transformed inputs to several homogeneous regions with mean filters on the spatial dimension to obtain smaller $K$ and $V$. SPRLT-Net \cite{Xue2022partition} divided the patch features into smaller overlapping sub-patches and obtained $Q$, $K$, and $V$ based on sub-patches rather than each element in the patch feature. HFTNet \cite{Yadav2024HFTNet} proposed to divide the $Q$ vector into two parts, one of which was processed by self-attention, while the other was processed by convolutional layers fused through residual connection. SSPT \cite{Ma2023tokenreduction} only kept the top k tokens with high attention scores, and SSBFNet \cite{Wu2024ssbfnet} kept the top k highly relevant key-value pairs by matrix multiplication between $Q$ and transpose of $K$. FTSCN \cite{Liang2023SimAMCNN} generated axial features horizontally and vertically, output different $Q$, $K$, and $V$ concerning the different axial features, calculated self-attention separately and then added the two attentions together. MaskedSST \cite{Linus2023marsked} also adopted a similar approach by sequentially processing the spatial and spectral feature tokens. LRDTN \cite{Ding2024LRDTN} used convolution layers with a kernel size of 1$\times$1 to reduce the spatial dimension of the input sequence before converting the feature map into $Q$, $K$, and $V$. 
	
	Moreover, because the channel dimension, which can be further divided, is usually smaller than the spatial dimension, CDSFT \cite{Qiu2023cdsft} proposed to decrease the computation complexity by transposing the input features so that the self-attention computation complexity is quadratic to the spectral dimension rather than the spatial dimension. ELViT \cite{Tan2024elvit} adopted a linear vision Transformer, which changes the left self-attention multiplication to right multiplication. 
	
	As shown in Fig. \ref{els}, ELS2T \cite{Zhang2023ELS2T} used softmax on $Q$ to output context score, and the self-attention score for each token was only calculated concerning a potential token as:
	
	\begin{equation}
		\text {Attention}(Q, K, V, S)=\sum\left(\text {Softmax}\left( Q \right)K \right) \text{ReLU} \left( V \right)S
	\end{equation}
	
	\noindent where $S$ is the spectral feature through global pooling from the spectral branch. 
	
	LSGA-VIT \cite{Ma2023gaussian} used the original input feature $X$ instead of $K$ and $V$ to reduce the computation and parameters, as shown in Fig. \ref{lgsa}. With the 2D Gaussian function to model the spatial relationship, the attention formula became:
	
	\begin{equation}
		\text {Attention}(Q, X, X)=\text {Softmax}\left(\frac{QX^T}{\sqrt{d}} + {\mathbf{G}}(x,y) \right)X
	\end{equation}
	
	\noindent where the Gaussian function is given as:
	
	\begin{equation}
		{\mathbf{G}}(x,y) = {e^{ - \frac {{(x - t/2)^{2} + {(y - t/2)^{2}}}}{{2{\sigma ^{2}}}}}}
	\end{equation}
	
	\noindent where $t = h \times w$, $\sigma$ is the standard deviation, and $(x, y)$ represents the spatial position coordinates. This Gaussian absolute position encoding was also adopted by CAF-Former \cite{Xu2024CAF-Former}. Similarly, MATA \cite{Liu2023multiarea} only contained value and key without $Q$, as shown in Fig. \ref{mata}: 
	
	\begin{equation}
		\text {Attention}(K, V)=\text {Softmax}\left(\frac{KK^T}{\sqrt{d}}\right)V
	\end{equation}
	
	Furthermore, SGHViT \cite{rs15133366} replaced the self-attention module with three layers of 2D group convolutions, which can reduce the model parameters. It also replaced the feedforward network (FFN) with a conv feedforward network (CFFN), which contained two layers of 1$\times$1 2D convolutions. GSC-ViT \cite{Zhao2024groupwise} employed a group-wise separable multi-head self-attention (GSSA) module, partitioning the feature map along the spectral dimension into multiple groups and executing multi-head self-attention independently for each group. GRetNet \cite{Han2024gretnet} computed self-attention on a regional level and proposed Gaussian multi-head attention (GMA) to concentrate on the distinct spectral patterns across different heads. All these methods can decrease the computational complexity of MHSA on HSI data to different extents. However, the quadratic nature of the self-attention mechanism is still challenging to address while utilizing three-dimensional HSI data as input.

	\section{Skip connections}\label{sc}
	
	For a deep learning model, the skip connections are effective in reducing information loss, stabilizing the gradient propagation, enhancing the information exchange between network structures, and increasing the transmission of the network to the deep layers to alleviate the problem of gradient vanishing/explosion \cite{Xue2022partition, Xu2023ss1dswin}. In practice, the skip connection technique was widely utilized to develop networks for HSI classification tasks. It was not only used to transfer information between network modules but also used within various modules. 
	
	One type of shortcut connection is \textbf{Dense connection}, where each layer in the network has a connection to the previous layer \cite{rs13030498}. Assuming a deep learning network $H_L(.)$ with $L$ layers, there are $\frac{L(L+1)}{2}$ dense connections between each layer \cite{Pan2023dualtrans, Wu2022densely, Zu2023cascaded, Xu2023cs2dt}. 
	
	Moreover, SpectralFormer \cite{Hong2022spectralformer} stated that the short skip connection's information 'memory' ability was limited, while the long skip connection obtained insufficient fusion due to long-range information loss. Thus \textbf{Mid-range connection} was devised in SpectralFormer to learn cross-layer feature fusion adaptively, as shown in Fig. \ref{spectralformer}. This connection scheme was also utilized in other works \cite{He2022HyperViTGan, Tang2023double}. 
	
	MATNet \cite{Zhang2023MATNeT} proposed a flexible cross-layer connection method called \textbf{Multilayer Dense Connection} to choose suitable numbers of shallow, deep, and middle information to fuse according to the different needs of the task. This method might be more appropriate for the shortcut connection design of the diversified network architectures. In addition, ALSST \cite{rs16111912} integrated a learnable matrix, which was multiplied with the output of each residual block, in the Transformer encoder to diversify the feature tokens in deep models.

	\section{Loss function}\label{loss}
	
	For a supervised classification task, \textbf{Cross-Entropy Loss} is the most commonly adopted loss function. Given a total number of $N$ samples from $C$ categories, sample's label as $y$, and the corresponding predicted label as $\hat{y}$, the cross-entropy loss function is:
 	
 	\begin{equation} 
 		\text {Loss}_{ce} = -\frac {1}{N}\sum _{i=1}^{N}\sum _{j=1}^{C}\mathbf {y}_{i}^{j}\log {\hat {\mathbf {y}}_{i}^{j}}
 	\end{equation}
 	
 	\noindent which can also be written as the negative log likelihood loss as \cite{Liao2023fusion}:
 	
	\begin{equation}
		\text {Loss}_{ce}= -\frac {1}{N}\sum _{i=1}^{N}\sum _{j=1}^{C} {\left [{ {\mathbf {y}_{i}^{j} \log \left ({{\hat {\mathbf {y}}_{i}^{j}} }\right) + \left ({{1-\mathbf {y}_{i}^{j}} }\right)\log \left ({{1-\hat {\mathbf {y}}_{i}^{j}} }\right)}}\right]}
	\end{equation}
	
	GBiLSTM-MFCT \cite{Xu2022biLSTM} added a regularized loss to the cross-entropy loss by a weighting factor $\lambda$
	
	\begin{equation}
		\begin{aligned} 
			\text {Loss} =&\text {Loss}_{ce}+\lambda \text {Loss}_{o} \\
			=&-\frac {1}{N}\sum _{i=1}^{N}\sum _{j=1}^{C}\mathbf {y}_{i}^{j}\log {\hat {\mathbf {y}}_{i}^{j}}+\lambda \left \|{ \mathbf {W}''\mathbf {W}^{''^{T}} -\mathbf {I} }\right \|_{F}
		\end{aligned}
	\end{equation}

	\noindent where $\mathbf {W}''$ is basis vectors of convolution projector's weights, which was utilized to output feature map to calculate the loss function, $\mathbf {I}$ is identity matrix, and $\left \|\cdot\right \|_{F}$ is the Frobenius norm of a matrix. 
	
	In order to impose the classifiers to pay more attention to small-class samples and hard-to-classify samples, \textbf{Focal loss} was implemented \cite{rs14153705}. The Focal loss is a variant of cross entropy loss to improve classification performance by increasing the loss weight of small-class samples and difficult-to-classify samples as: 
	
	\begin{equation} 
		\text {Loss}_{c} = -\frac {1}{N}\sum _{i=1}^{N}\sum _{j=1}^{C} \alpha_{i} (1-p_{i}^{j})^{\gamma} \mathbf {y}_{i}^{j}\log {\hat {\mathbf {y}}_{i}^{j}}
	\end{equation}
	
	\noindent where $\alpha\in\mathbb{R}^{1\times N}$ is a weighting factor and $(1-p)^{\gamma}$ is a tunning factor with $\gamma\geq0$. LSDnet \cite{Peng2024LSDnet} proposed to dynamically adjust $\gamma$ for different epochs as $\gamma = \frac {\alpha }{\sqrt {\beta }}$, where $\beta$ denotes the number of epochs for the current training. In another work, SSACT \cite{rs15061612} added the class intersection over union (Class-IoU) $\text{Class}_{IoU} = \frac{TP}{TP + FP + FN}$ into the Cross-Entropy loss function to diminish the inter-class classification difference caused by the class imbalance of samples. 
	
	Moreover, for the cross-entropy loss function, only the loss of the correct label is considered, while the loss of other labels is ignored \cite{rs13030498}. Therefore, \textbf{Label smoothing} was adopted to mitigate the overfitting problem and increase the generalization ability of the model by assigning tiny values on the other label positions \cite{Ding2023crossdomain}. Given $y_c$ as the one-hot representation of each label, whose dimension is $C$, as the number of classes, the value of label position is 1 and 0 otherwise. The smoothing factor (small noise value) to the label as \cite{rs13030498, Dang2023DoublebranchFF}:
	
	\begin{equation}
		y_{c}^{^{\prime}} = \left( {1 - \varepsilon } \right)y_{c} + \frac{\varepsilon }{C}
	\end{equation}
	
	\noindent where ${y}_{n}^{^{\prime}}$ is the new label. Therefore, the label smoothing cross-entropy (LSCE) loss can be obtained as \cite{Zhang2023MATNeT}:
	
	\begin{equation}
		\text {Loss}_{LSCE} = \left( {1 - \varepsilon } \right) \text {Loss}_{m} + \varepsilon \sum \frac{ \text {Loss}_{n} }{C}
	\end{equation}
	
	\noindent where $\text {Loss}_{m}$ represents the standard cross-entropy loss of correct class $m$, and $n$ is the incorrect class. In addition, a \textbf{Label Smoothing Poly (Lpoly)} was proposed to dynamically change the smoothness of different prediction labels by reducing the weight of the correct labels \cite{Zhang2023MATNeT}: 
	
	\begin{equation}
		\text {Loss}_{Lpoly} = \left( {1 - \varepsilon } \right) \text {Loss}_{m} + \varepsilon \sum \frac{ \text {Loss}_{n} }{C} + \gamma\left( 1 - \hat {\mathbf {y}}_{i}^{j} \right)
	\end{equation}
	
	\noindent where$\hat {\mathbf {y}}_{i}^{j}$ represents the predicted probability that the observed sample $i$ belongs to class $j$. Moreover, the Polyloss \cite{Qi2023globallocal} is formulated as: 
	
	\begin{equation}
		\text {Loss}_{poly} = \text {Loss}_{ce} + \gamma\left( 1 - \hat {\mathbf {y}}_{i}^{j} \right)
	\end{equation}
	
	Additionally, DCN-T \cite{Wang2023DCN-T} adopted a soft voting mechanism to take into account the probabilities that are neglected from the incorrect classes. HSI-TransUNet \cite{NIU2022107297} used a hybrid loss of cross-entropy loss and a region-based Log-Cosh Dice Loss, which is an improved version of \textbf{Dice loss}, and is calculated as:
	
	\begin{equation}
		\text {Loss}_{Dice} = \sum_{i=1}^{N}\left( 1 - \frac{2|\mathbf {y}_{i} \bigcap \hat {\mathbf {y}}_{i}| + \varepsilon}{|\mathbf {y}_{i}| + |\hat {\mathbf {y}}| + \varepsilon}\right)
	\end{equation}
	
	\noindent and given $cosh(x) = \left(\frac{e^x+e^{-x}}{2}\right)$, the proposed Log-Cosh Dice loss is $\text {Loss}_{lc-dce} = log\left(cosh\left(\text {Loss}_{Dice}\right)\right)$. Thus the hybrid loss is given by:
	
	\begin{equation}
		\text {Loss} = \left(1-\alpha\right) \text {Loss}_{ce} + \alpha \text {Loss}_{lc-dce}
	\end{equation}
	
	Numerous works utilized the \textbf{Joint Loss} as a loss function because it is flexible to adjust the importance of different features or modules with fixed or learnable weights \cite{Zhou2022hussat}. There are a variety of different losses joined using this method, including the cross-entropy loss, task-specific reconstruction loss, with compactness loss \cite{Zou2022LessFormer}, reconstruction loss with contrastive loss \cite{Cao2023maskedAECL}, cross-entropy loss, the supervised contrastive loss, the unsupervised contrastive loss, with the regularization term \cite{Zhou2023ViTContrast}, classification loss with mask prediction loss \cite{Bai2022multibranch}, cross-entropy loss for main and auxiliary loss \cite{Wang2023DCN-T}, dice loss with focal loss \cite{rs15143491}, and contrastive loss with conditional domain discriminating loss \cite{Liu2024stdbip}. 
	
	The loss functions can also be used to fuse results of different granularities with the adaptive weighted fusion method \cite{Ouyang2023hybrid} and spatial-spectral features from different branches \cite{Xin2022conhsi}. Moreover, HSDBIN \cite{Qin2024HSDBIN} proposed to fuse uniformity loss based on a Gaussian latent kernel and a structural-aware distillation loss, which measures the similarity between the hyperspherical space and the metric space of the classifier's output logits. 
	
	The \textbf{Joint Loss} method is a flexible and straightforward approach to aggregating the outputs of different modules. However, using adaptive factors and, in addition to linearly integrating the output of different branches in the loss function, may not fully capture the non-linear relationship between feature maps.

	\section{Future research directions}
	
	Although numerous research papers have been devoted to designing Transformer-based models for HSI classification tasks and demonstrated promising results, some challenges are still worth further exploration and research.
	
	\begin{enumerate}
		\item \textbf{Limited hyperspectral datasets}: Obtaining new labeled hyperspectral datasets is costly. Thus, most of the research papers surveyed in this work used publicly available datasets, including Indian Pines, Pavia University, Salinas, and Houston, as shown in Table \ref{data}. The lack of data is even more severe in specific areas such as food quality control, crop mapping, and medical imaging analysis. The limited data issue hinders the research from fully understanding the types and extent of problems and developing appropriate solutions. Therefore, more datasets covering different areas should be collected, and the algorithms to address the limited sample issue should be further developed. 
		
		\item \textbf{HSI system size and cost}: The hyperspectral imaging system is relatively larger and more expensive than the RGB one. These issues hamper the development of compact systems and limit the application scopes of HSI systems. Although some methods have already been proposed to improve HSI systems to be less bulky \cite{Lin2023metasurface}, more research should be devoted to developing more affordable and compact HSI systems.
		
		\item \textbf{Computational cost}: The computational complexity of multi-head self-attention is quadratic to the input tokens, and the data size of the HSI datacube is tens or even hundreds of times larger than the RGB images with the same pixel number. Combining Transformer-based models with HSI produces a heavy computational burden and substantial memory requirements. Moreover, the cascaded or parallel network architectures consume even more memory space to extract spatial-spectral features better. This issue is a significant obstacle to deploying Transformer-based networks for HSI application systems, such as portable devices for agriculture contamination detection, food product quality control, or embedded systems in UAVs for agriculture monitoring.

		\item \textbf{Robustness and generalization}: HSI is susceptible to multiple variations and interferences, such as light intensities, light reflection angles, and temperature. Future research could focus on developing more advanced techniques to handle these challenges more efficiently and effectively for applications on a wide range of datasets under different situations.
		
		\item \textbf{Explainability}: Transformer-based models are treated as black boxes, which lack transparency and interpretability. Besides the research that focused specifically on improving the explainability of self-attention mechanism \cite{Chefer_2021_CVPR} and generic Transformer architectures \cite{Chefer_2021_ICCV}, there have also been some studies that integrated general Explainable Artificial Intelligence (XAI) methods with Transformer, such as land use classification with Integrated Gradients \cite{Khan2024xailulc}, Leaf disease severity classification with Grad-CAM \cite{bandi2023leaf}, chest X-ray diagnosis with Grad-CAM and Information Bottleneck Attribution (IBA) techniques \cite{Demir2024medicaxai}, and incident heart failure prediction with feature perturbation method \cite{Rao2022xai}. Some researchers also reviewed the XAI methods applied for Transformers \cite{computers13040092xai, kashefi2023xaitrans} and evaluated the effects of different XAI algorithms \cite{electronics13010175xai}. However, no method has been specifically designated to explain the relevance of the spatial-spectral feature of HSI with transformer-based models. In addition, explainability and trustworthiness are essential requirements for many real-world decision-making systems. Therefore, developing more explainable and interpretable Transformer-based networks for HSI applications, which can provide more insightful information about the intrinsic relationships within HSI data, is another important direction for future research.

	\end{enumerate}

	\section{Conclusion}
	This research study thoroughly explores how Transformer-based models are used in classifying imaging (HSI), displaying the growing connection between advanced deep learning techniques and spectral imaging. With a wealth of studies utilizing Transformers for analyzing HSI data and a lack of reviews on the topic, this study delves into the latest advancements to overcome the challenges of processing HSI data using Transformer architectures. It systematically discusses the stages and approaches employed in Transformer-based models for HSI classification, addressing issues like small sample sizes, token embedding, spatial-spectral feature extraction, integration of multi-head self-attention mechanisms, and optimizing loss functions. This detailed overview assists researchers in improving and innovating components of Transformer-based models and developing new learning strategies. 
	\\
	\\
	\textit{CRediT authorship contribution statement}\\
	\textbf{Guyang Zhang}: Conceptualization, Methodology, Investigation, Writing- Original draft, Visualization, Writing- Reviewing and Editing. \textbf{Waleed Abdulla}: Conceptualization, Project administration, Resources, Visualization, Supervision, Writing- Reviewing and Editing.

	\newpage
	
	\medskip

	\begin{appendices}
		\section{Reviewed paper summary}
		
		\begin{longtable}[H]{|p{1.5cm} | p{15cm}|}
			\caption{Paper summary for data bases \label{databases}} \\
			\hline
			\multicolumn{2}{|c|}{ Paper Summary} \\
			\hline
			Database & Transformer Methods \\
			\hline
			\endhead
			IEEE 2020 - 2022 & HSI-BERT \cite{He2020hsibert}, STransFuse \cite{Gao2021stransfuse}, 3DSwinT \cite{Huang2022swinvit}, HSI-TNT \cite{Liu2022coastal}, SpectralFormer \cite{Hong2022spectralformer}, CTMixer \cite{Zhang2022mixer}, Central Attention Network (CAN) \cite{Liu2022central}, HiT \cite{Yang2022hsiT}, CTN \cite{Zhao2022contrans}, SSFTT \cite{Sun2022tokenization}, SPRLT-Net \cite{Xue2022partition}, GAHT \cite{Mei2022groupaware}, HyperViT \cite{Praven2022HyperViT}, MSTNet \cite{Yu2022mstnet}, BS2T \cite{Song2022bs2t}, SST-M \cite{Bai2022multibranch}, LESSFormer \cite{Zou2022LessFormer}, LSFAT \cite{Tu2022aggregation}, CASST \cite{Peng2022crossattention}, GBiLSTM-MFCT \cite{Xu2022biLSTM}, CSiT \cite{He2022CSiT}, HSST \cite{Song2022hierarchical}, Spatial Sample Selection(3S) \cite{Feng2022gaussian}, BERT-HyperSLIC-DBSCAN \cite{Sigirci2022hyperslic}, Spa-Spe-TR \cite{He2022twobranch}, \\
			\hline
			IEEE 2023 & CTFSN \cite{Zhao2023fusionsplicing}, DATE \cite{Tang2023double}, S2FTNet \cite{Liao2023fusion}, RNN–Transformer (RT) \cite{Zhao2023rnntrans}, morphFormer \cite{Roy2023morpho}, MAR-LWFormer \cite{Fang2023multiattention}, CDCformer \cite{Zu2023cascaded}, DCN-T \cite{Wang2023DCN-T}, GTCT \cite{Qi2023globallocal}, GMA-Net \cite{Lu2023multiattention}, ITCNet \cite{Yang2023multilevelIT}, MRViT \cite{Cao2022mixedres}, SS1DSwin \cite{Xu2023ss1dswin},  Spectral-MSA \cite{Zhou2021swinspectral}, Brain-tissue \cite{Cruz-Guerrero2023braintissues}, CESSUT \cite{Xin2022conhsi}, ELS2T \cite{Zhang2023ELS2T}, FTSCN \cite{Liang2023SimAMCNN}, HUSSAT \cite{Zhou2022hussat}, Co-learning \cite{Chen2022colearning}, HCVN \cite{Yan2023hybridconvvit}, LSGA \cite{Ma2023gaussian}, MATNet \cite{Zhang2023MATNeT}, MethaneMapper \cite{Kumar2023methanemapper}, MCAL \cite{Xu2023multiscale}, spatial-spectral-based 3D ViT \cite{Zhou2022rethinking}, SEDT \cite{Wu2022densely}, MSTNet \cite{Yu2022mstnet}, MSVT \cite{Chen2021multistageViT}, FUST \cite{Zeng2023microscopic}, 
			SSTF-Unet \cite{Liu2023sstfunet}, TMAC \cite{Cao2023maskedAECL} MSDFormer \cite{Chen2023msdformer}, MSNAT \cite{Qian2023neighborhood}, SSTFSL \cite{Cao2023crossdomain}, Grid-Transformer \cite{Guo2023gridfewshot}, MATA \cite{Liu2023multiarea}, TransCNN \cite{Hao2022featurefusion}, SSTE-Former \cite{Wu2023tokenhash}, FFTN \cite{Liu2022feedback}, MSBMSW \cite{Li2023shift}, MaskedSST \cite{Linus2023marsked}, CTFSL \cite{Peng2023crossdomain}, CAL \cite{Wang2023improvedCapsule}, ToMF-B \cite{Zhao2022crossscene}, HyperViTGAN \cite{He2022HyperViTGan}, SSTNet \cite{Kang2023ssloil}, SS-MTr \cite{Huang2023markedtrans}, SITS \cite{Yuan2021pretrainvit}, MHCFormer \cite{Shi2023mhcformer}, FactoFormer \cite{Mohamed2023factorformer}, SANet \cite{Zhang2023SAnet}, IMAE \cite{Kong2023instoken}, ITER \cite{Yang2023iter}, Cross-domain calibration \cite{Ding2023crossdomain}, SPFormer \cite{Li2023spformer}, CMTL \cite{Cheng2023causalmeta}, MSSFP \cite{Zhou2023MSSFP}, MST-SSSNet \cite{Gao2023MSTSSSNet}, D2S2BoT \cite{Zhang2023D2S2BoT}, LGGNet \cite{Zhang2023LGGNet}, GAB-UFCN \cite{Zhou2023GABUFCN}, DFTN \cite{Qiao2023fdfe}, SpecTr \cite{Yun2023SpecTr}, CD-ViT \cite{Ling2023cdvit}, CDSFT \cite{Qiu2023cdsft}, PUSL \cite{Yao2023pusl}, PASSNet \cite{Ji2023passnet}, TRUG \cite{Hao2023trug}, CS2DT \cite{Xu2023cs2dt}, G2T \cite{Shi2023g2t}, Concrete Crack Segmentation \cite{Steiner2023concrete}, SSFT \cite{Qiao2023sstf}, MIEPN \cite{Yang2023MIEPN}, GTFN \cite{Yang2023gtfn}, \cite{Viel2023analysis}, GSPFormer \cite{Chen2023projection}, SSPT \cite{Ma2023tokenreduction}, CBFF-Net \cite{Gao2023cbffNet}, SLA-NET \cite{Zhang2023morpho}, LoFTR \cite{Perera2023lowpixel}, HSIC-FM \cite{Yang2023barrier}, Tinto \cite{Afifi2023tinto}, DAFFN \cite{Cui2023DAFFN}, TMAC \cite{Cao2023maskedAECL}, LiT \cite{Zhang2023lightweight}, CViT \cite{Zhou2023ViTContrast}, HybridFormer \cite{Ouyang2023hybrid}, SSLSM \cite{Liu2023sslmask}, \cite{Li2023generative}, MAT-ASSAL \cite{Zhao2023multiattention}   \\
			\hline
			IEEE 2024 & Hierarchical attention Transformer \cite{Arshad2024hierarchi}, WaveFormer \cite{Ahmad2024waveformer}, CCSF-Transformer \cite{Zhu2024centercate}, GSC-VIT \cite{Zhao2024groupwise}, HMSSF \cite{He2024hybridmultiscale}, DBMST \cite{Shi2024dbmst}, LSDnet \cite{Peng2024LSDnet}, SWFormer \cite{Li2024swformer}, H2MWSTNet \cite{Zhong2024h2mwstnet}, Dual-MTr \cite{Li2024DualMTr}, MCTT \cite{Wang2024MCTT}, MASSFormer \cite{Sun2024massformer}, GraphGST \cite{Jiang2024GraphGST}, MST-SSSNet \cite{Gao2024MST-SSSNet}, MHIAIFormer \cite{Kong2024MHIAIFormer}, CAMFT \cite{Yang2024CAMFT}, LRDTN \cite{Ding2024LRDTN}, SVAFormer \cite{Chen2024svaformer}, CAT \cite{Feng2024cat}, SQSFormer \cite{Chen2024SQSFormer}, SMESC \cite{Yu2024SMESC}, SS-VFMT \cite{Huang2024SS-VFMT}, MSMT-LCL \cite{Zhou2024msmt-lcl}, S$^2$GFormer \cite{Huang2024S2GFormer}, Cross-Dataset \cite{Bai2022multibranch}, CAF-Former \cite{Xu2024CAF-Former}, SPTNet \cite{Ma2024SPTNet}, CMT \cite{Jia2024CMT}, CentralFormer \cite{Li2024CentralFormer}, RMAE \cite{Wang2024RMAE}, DCTN \cite{Zhou2024dctn}, MS2I2Former \cite{Cheng2024MS2I2Former}, DNAT \cite{Tejasree2024dnat}, CACFTNet \cite{Cheng2024CACFTNet}, PyFormer \cite{Ahmad2024PyFormer}, U2ConvFormer \cite{Zhan2024U2ConvFormer}, GRetNet \cite{Han2024gretnet}, CSJA \cite{Li2024csja}, SCM-CT \cite{Zhao2024SCM-CT}, CTF-SSCL \cite{Xi2024CTF-SSCL}, DISGT \cite{Cheng2024disgt}, HSDBIN \cite{Qin2024HSDBIN}, EHSnet \cite{Wang2024EHSnet}, CreatingNet \cite{Li2024CreatingNet}, DAHIT \cite{Shi2024DAHIT}, DBSSAN \cite{Zhao2024dbssan}, CD-DViT \cite{Ye2024CD-DViT}, SRT \cite{Zhou2024srt}, DEMAE \cite{Li2024demae}, HyperSINet \cite{Yu2024hypersinet}, APSFFT \cite{Huang2024APSFFT}, MSSTT \cite{Meng2024msstt}, Cross-datasets \cite{Bai2024crossdata}    \\
			\hline
			Elsevier & Pests and disease classification \cite{LIU2022107448}, maize damage detection \cite{LIU2023107853}, HSI-TransUNet \cite{NIU2022107297}, MCE-ST \cite{KHOTIMAH2023103286}, CSIL \cite{YANG2023145}, LAGAN \cite{CHEN2023120828}, Hyper-ES2T \cite{WANG2022103005}, RDNT \cite{LI2024123939}, TECCNet \cite{PAN2024106973}, NeiCoT \cite{LIANG2024103979}, DATN \cite{SHU2024107351}, EggFormer \cite{JI2024109298}, Fire detection \cite{YANG2024105104}   \\
			\hline
			Springer & Double-branch feature fusion Transformer \cite{Dang2023DoublebranchFF}, small-data convex/deep (CODE) \cite{Lin2023metasurface}, CST \cite{Cai2022cst}, DKAT \cite{Tejasree2023}, DiCT \cite{Zhou2023dictionary}, MT-CW \cite{Tulapurkar2022mha}, maiz seeds \cite{Domezz2023hybrid}, ELViT \cite{Tan2024elvit}, DiffSpectralNet \cite{Sigger2024DiffSpectralNet}, HFTNet \cite{Yadav2024HFTNet}, STBDIP \cite{Liu2024stdbip}, SSBFNet \cite{Wu2024ssbfnet},  \\
			\hline
			Wiley & CITNet \cite{Liao2023integrated}, \cite{Zhou2023change}, DT-FSL \cite{Ran2023fewshot}, S3FFT \cite{Xie2023fusion}, Bacteria \cite{Lu2024bacteria}, MedDiffHSI \cite{Sigger2024brain} \\
			\hline
			MDPI & FusionNet \cite{rs14164066}, CAEVT \cite{s22103902}, NEHT \cite{rs14194732}, IFormer \cite{rs14194866}, S2Former \cite{electronics12183937}, SST \cite{rs13030498}, MFSwin-Transformer \cite{agronomy12081843}, Enhanced TabNet \cite{rs14030716}, AttentionHSI \cite{rs14091968}, MSSTT \cite{electronics12183879}, HyperSFormer \cite{rs15143491}, Spatial Shuffle \cite{rs15163960}, SGHViT \cite{rs15133366}, SCSTIN \cite{rs15133269}, SS-TMNet \cite{rs15051206}, SDFE \cite{rs15010261}, CTAFNet \cite{app13010492}, \cite{rs14143426}, SSACT \cite{rs15061612}, HSD2Former \cite{rs16234411}, RS-Net \cite{electronics13204046},  Glaucoma Detection \cite{diagnostics14121285}, ALSST \cite{rs16111912}, TNCCA \cite{rs16071180}, Cherry Tomatoes \cite{foods13020251}, MSST \cite{rs16020404}, D$^2$BERT \cite{rs16030539}, DCTransformer \cite{app14051701}, S3L \cite{rs16060970},  \\
			\hline
			Taylor \& Francis & DTT-TRM \cite{Liu2022dsstrm}, SpectralSWIN \cite{Selen2022SpectralSWIN}, RSAGformer \cite{Zu2023wrsag}, CTIN \cite{Li2023cnntrinter}, MSTViT \cite{Li2022granu}, MDvT \cite{Zhou2023mdvt}, MS$^3$DT \cite{Pan2023dualtrans}, MLFF \cite{Hao2023multilayer}, HybridGT \cite{Neela01022025}, Grassland \cite{Zhang18032024} \\
			\hline

		\end{longtable}
		
		\newpage

		\section{Datasets}
		
		\begin{longtable}[H]{|p{5cm} | p{13cm}|}
			\caption{Summary for public datasets \label{data}} \\
			\hline
			\multicolumn{2}{|c|}{ Public Datasets} \\
			\hline
			Datasets & Links \\
			\hline
			\endhead
			IP, PU, PC, SA, KSC, Botswana, Cuprite & \url{https://www.ehu.eus/ccwintco/index.php?title=Hyperspectral_Remote_Sensing_Scenes}  \\
			\hline
			The Houston2013 & \url{http://www.grss-ieee.org/community/technical-committees/data-fusion/2013-ieee-grss-data-fusion-contest/} \\
			\hline
			Houston 2018 & \url{https://hyperspectral.ee.uh.edu/?page_id=1075} \\
			\hline
			EnMAP, EnMAP-DFC \cite{Linus2023marsked} &  \url{https://github.com/HSG-AIML/MaskedSST} \\
			\hline
			WHU & \url{http://rsidea.whu.edu.cn/resource_WHUHi_sharing.htm} \\
			\hline
			Chikusei \cite{NYokoya2016} & \url{https://www.sal.t.u-tokyo.ac.jp/hyperdata/} \\
			\hline
			HyRANK & \url{https://zenodo.org/records/1222202} \\
			\hline
			ZY1-02D & \url{http://sasclouds.com/chinese/normal/}, 
			\\
			\hline
			Xiongan (XA) datasets & \url{http://www.hrs-cas.com/a/share/shujuchanpin/2019/0501/1049.html} \\
			\hline
			AeroRIT dataset & \url{https://github.com/aneesh3108/AeroRIT} \\
			\hline
			Tea farm & \url{https://doi.org/10.3974/geodb.2017.03.04.V1} \\
			\hline
			the University of Southern Mississippi Gulfpark (MUUFL) & \url{https://github.com/GatorSense/MUUFLGulfport/tree/master} \\
			\hline
			Xuzhou dataset & \url{https://ieee-dataport.org/documents/xuzhou-hyspex-dataset} \\
			\hline
			MethaneMapper & https://github.com/UCSB-Vrl/MethaneMapper-Spectral-Absorption-aware- Hyperspectral-Transformer-for-Methane-Detection. \\
			\hline
			Salt stress detection in wheat & \url{https://conservancy.umn.edu/handle/11299/195720} \\
			\hline
			MHSI Choledoch Dataset & https://www.kaggle.com/datasets/hfutybx/mhsi-choledoch-dataset-preprocessed-dataset?resource=download \\
			\hline
			Multidimensional Choledoch Database \cite{Zhang2019Cholangiocarcinoma} & http://bio-hsi.ecnu.edu.cn \\
			\hline
			Oil spill benchmark database (HOSD) over Gulf of Mexico & \url{https://drive.google.com/file/d/1MKBcASK22931kqsUT886n7Ufdz3g_GZ8/view?usp=sharing} \\
			\hline
			Vaihingen and Postdam & \url{https://www2.isprs.org/commissions/comm2/wg4/benchmark/2d-sem-label-vaihingen/},  \url{http://www2.isprs.org/commissions/comm3/wg4/2d-sem-label-potsdam.html} \\
			\hline
			Xinjiang Cotton & \url{https://zenodo.org/records/7856467} \\
			\hline 
			In-Vivo Hyperspectral Human Brain Dataset \cite{Leon2023brain}  &  \url{https://hsibraindatabase.iuma.ulpgc.es} \\
			\hline
			Oral cancer histopathology dataset & \url{https://www.kaggle.com/datasets/ashenafifasilkebede/dataset} \\
			\hline

		\end{longtable}

	\end{appendices}


\end{document}